\documentclass[acmsmall]{acmart}
\AtBeginDocument{%
  }

\setcopyright{acmlicensed}
\copyrightyear{2025}
\acmYear{2025}
\acmDOI{XXXXXXX.XXXXXXX}

\acmJournal{CSUR}

\usepackage{subfigure}
\usepackage{array}
\newcolumntype{P}[1]{>{\centering\arraybackslash}p{#1}}
\usepackage{bm}

\DeclareMathOperator*{\argmax}{arg\,max}

\begin{document}

\title{Sustainable Smart Farm Networks: Enhancing Resilience and Efficiency with Decision Theory-Guided Deep Reinforcement Learning}

\author{Dian Chen}
\email{dianc@vt.edu}
\affiliation{%
  \institution{Virginia Tech}
  \city{Arlington}
  \state{VA}
  \country{USA}
}

\author{Zelin Wan}
\email{zelin@vt.edu}
\affiliation{%
  \institution{Virginia Tech}
  \city{Arlington}
  \state{VA}
  \country{USA}
}

\author{Dong Sam Ha}
\email{dha@vt.edu}
\affiliation{%
  \institution{Virginia Tech}
  \city{Arlington}
  \state{VA}
  \country{USA}
}

\author{Jin-Hee Cho}
\email{jicho@vt.edu}
\affiliation{%
  \institution{Virginia Tech}
  \city{Arlington}
  \state{VA}
  \country{USA}
}

\begin{abstract}
  Solar sensor-based monitoring systems have become a crucial agricultural innovation, advancing farm management and animal welfare through integrating sensor technology, Internet-of-Things, and edge and cloud computing. However, the resilience of these systems to cyber-attacks and their adaptability to dynamic and constrained energy supplies remain largely unexplored. To address these challenges, we propose a sustainable smart farm network designed to maintain high-quality animal monitoring under various cyber and adversarial threats, as well as fluctuating energy conditions. Our approach utilizes deep reinforcement learning (DRL) to devise optimal policies that maximize both monitoring effectiveness and energy efficiency. To overcome DRL's inherent challenge of slow convergence, we integrate transfer learning (TL) and decision theory (DT) to accelerate the learning process. By incorporating DT-guided strategies, we optimize monitoring quality and energy sustainability, significantly reducing training time while achieving comparable performance rewards. Our experimental results prove that DT-guided DRL outperforms TL-enhanced DRL models, improving system performance and reducing training runtime by 47.5\%.
\end{abstract}




\maketitle

\section{Introduction} \label{sec:intro}

\subsection{Motivation \& Goal} \label{subsec:motivation-goal}
Solar sensor-based animal monitoring systems are pivotal, energy-efficient innovations that minimize operational costs in agriculture. These systems leverage advancements in sensor technology, the Internet of Things (IoT), edge computing, and cloud computing to enhance farm management and animal welfare~\cite{bogue2012solar}. In a smart farm, each animal, such as a cow, is outfitted with a small sensor attached to its ear to collect data on vital signs and behaviors, enabling remote monitoring. This data is transmitted to gateways using Long Range (LoRa) technologies and forwarded to a cloud server for analysis and storage. However, reliance on digital communication and network protocols introduces vulnerabilities. Compromised or missed data on animal conditions can undermine monitoring quality, impacting accurate disease prediction and other analyses. This highlights the need for improved convergence to ensure high-quality monitoring and optimal energy efficiency.

To tackle the challenges associated with sensor maintenance, such as the need for frequent battery replacements, we propose the adoption of solar-powered sensors. This approach eliminates the labor-intensive and expensive task of regularly changing batteries. However, the limited size of the solar panels imposes constraints on the total energy available to the sensors. Consequently, it becomes imperative to establish an optimal energy policy that extends the sensors' energy lifetime without compromising the efficacy of the monitoring system. 

{\bf The goal of this work} is to develop a monitoring system that is both resilient to attacks and adaptable to energy needs, specifically designed for smart farms operating under resource limitations.  Traditional approaches have primarily utilized rule-based systems in complex system management, especially under adversarial conditions~\cite{alemayehu17}. These systems, however, face significant challenges in addressing the complexities and dynamics inherent to such frameworks, particularly in adversarial settings. The rigidity of rule-based mechanisms often results in suboptimal performance under these conditions.  To overcome these challenges, we adopt a deep reinforcement learning (DRL) strategy~\cite{schu17}.  DRL integrates deep learning with reinforcement learning agents, enabling them to adjust behavior based on actions and rewards, achieving robust end-to-end control and significant progress in tasks with high-dimensional inputs~\cite{kai2017}.  We introduced this DRL approach to identify the best setting to enhance data monitoring quality while ensuring sensor networks' energy sustainability under adversarial threats. However, DRL has suffered from slow convergence, particularly in highly dynamic, uncertain, or adversarial environments.  Further, DRL faces challenges in learning efficiency due to costs associated with real-world interactions. Despite its trial-and-error learning process, the initial performance of DRL agents can be inferior to non-learning approaches.

In addition, to expedite the learning curve, we integrate transfer learning (TL) with our DRL model, inspired by~\cite{zhu2023transfer}. This integration aims to utilize pre-existing models to speed up the learning process. However, the effectiveness of TL hinges on the availability of extensive datasets for pre-trained models, and a dearth of such data can lead to increased communication costs necessary for achieving the desired learning outcomes.  Despite DRL's capability to autonomously identify optimal solutions in volatile environments, its prolonged convergence time remains a significant hindrance. To tackle this issue, we propose a novel approach combining decision theory (DT) and DRL.  DT evaluates and compares actions based on their expected utility, considering probabilities and outcomes' values~\cite{peterson2017introduction, parmigiani2009decision}. While often outperforming DRL approaches in the early stage of simulations, DT can be impractical in real life due to uncertainties in value evaluation and preferences among risks. As environments grow more complex and dynamic, designing a utility function that accurately reflects preferences becomes challenging and time-consuming~\cite{north1968tutorial}. In addition, compared to DRL, DT is limited in finding optimal solutions and tends to get stuck in local optima. Thus, we designed this DT-guided DRL strategy to mitigate the limitations associated with applying individual DT, DRL, or TL in these contexts, mainly focusing on the protracted learning periods and the extensive pre-training often required by TL. By incorporating DT into the DRL framework, we aim to accelerate DRL's learning capability in managing complex and dynamic systems under cyber and adversarial attacks.

\subsection{Key Contributions} \label{subsec:key-contributions}
This work makes the following {\bf key contributions}: 
\begin{itemize}
    \item We introduce a solar sensor-based smart farm monitoring system resilient against cyber and adversarial attacks while adapting to energy needs. This system addresses gaps in existing research by focusing on such attacks in energy-limited environments \cite{metallidou2020energy, liu12}.
    \item We propose a novel approach, named {\em DT-guided DRL}, which integrates DT into DRL to address the prolonged convergence time of traditional DRL and the extensive pre-training requirements of TL-based DRL. By leveraging DT’s ability to evaluate actions based on expected utility, our method accelerates learning in complex, dynamic, and adversarial environments while maintaining DRL's capability to find global optima. This approach combines the strengths of DT and DRL, mitigating the limitations of each to enhance efficiency and adaptability in managing intricate systems.
    \item We tackle epistemic uncertainty in animal health monitoring using {\em Subjective Logic}~\cite{sj16}, contrasting with prior focus on aleatoric uncertainty~\cite{aerts2023bayesian}, to improve predictions of conditions like temperature and heartbeat frequency.
    \item We improve monitoring quality and energy management by integrating DRL, TL, and DT. This includes fine-tuning TL models and DT-guided DRL, offering a novel approach to smart farm research.
    \item We confirmed the efficacy of our approach via extensive simulations with semi-synthetic datasets. Fully fine-tuned TL models enhanced monitoring quality, while DT-guided DRL was most effective for energy management, demonstrating their promising applications in various settings.
\end{itemize}

\section{Related Work} \label{sec:related-work}

\subsection{Decision Theory \& Deep Reinforcement Learning}

Decision theory (DT) evaluates and compares actions based on their expected utility, considering probabilities and outcomes' values~\cite{peterson2017introduction, parmigiani2009decision}. While effective under ideal conditions, adhering to the decision-makers preferences, DT can be impractical in real life due to uncertainties in value evaluation and preferences among risks. As environments grow more complex, designing a utility function that accurately reflects preferences becomes challenging and time-consuming~\cite{north1968tutorial}.  

Deep reinforcement learning (DRL) integrates deep learning with reinforcement learning agents, enabling them to adjust behavior based on actions and rewards, achieving robust end-to-end control and significant progress in tasks with high-dimensional inputs \cite{kai2017}. However, DRL faces challenges in learning efficiency due to costs associated with real-world interactions. Despite its trial-and-error learning process, the initial performance of DRL agents can be inferior to non-learning approaches.

This work combines DT with deep reinforcement learning (DRL) to achieve close-to-optimal results and fast training. The human-designed DT component provides an initial decision strategy, ensuring the algorithm starts with acceptable performance rather than random actions. By integrating DT's utility distribution with DRL's neural networks, our approach avoids random beginnings and guides DRL training more effectively.

\subsection{Smart Farm Sensor Systems} \citet{kumar19} developed gCrop, an IoT-based system that employs machine learning and computer vision to assess and predict leaf growth, optimized for low-power usage in energy-sensitive environments.  \citet{liu12} introduced a meta-heuristic approach to enhance the operational efficiency of dynamic wireless sensor networks (WSNs). Contrasting with the efforts of~\cite{kumar19, liu12}, our research is centered on creating models that are resilient to attacks and robust against environmental uncertainties and dynamics, aiming to sustain high-quality monitoring and ensure energy sustainability within sensor networks.

In the realm of smart farming, managing constrained resources within wireless sensor networks (WSNs) is a pivotal challenge, with energy consumption and battery life playing crucial roles in the efficacy of real-time monitoring. \citet{alemayehu17} introduced an efficient heuristic algorithm for the traveling salesman problem to enhance data monitoring performance while minimizing computational costs. Notably, the balance between monitoring system effectiveness and efficiency, especially in smart sensor systems for agriculture, has seldom been addressed in existing research. Drawing inspiration from~\cite{alemayehu17}, we developed a rule-based heuristic for performance comparison, marking a novel approach in the study of smart farms.

Despite the limited focus on cybersecurity in existing smart system research, the advent of sophisticated cyberattacks has underscored the critical vulnerabilities within these systems.  \citet{gupta20} shed light on cybersecurity challenges, particularly data and network attacks, in smart agricultural environments. \citet{saheed21} developed a mechanism for detecting cyberattacks in Internet-of-Medical-Things (IoMT) environments, utilizing a bio-inspired optimization algorithm to enhance sensor data features for training a novel deep recurrent neural network (RNN) for precise attack detection. Furthermore, \citet{chae2018enhanced} proposed a Peer-to-Peer (P2P) based smart farm system designed to thwart attackers' efforts to control communication and data, introducing an effective authentication method to expedite operational processes. \citet{aliyu2023blockchain} and \citet{vangala2021smart} introduced blockchain-based smart farm systems to bolster security and privacy, enabling the detection and mitigation of security threats through blockchain transactions associated with smart farming activities. Although blockchain technologies are renowned for enhancing security in smart systems, their substantial computational demands and impact on performance render them unsuitable for resource-limited settings. 

As highlighted by \cite{gupta20, saheed21}, there is a notable gap in research addressing security issues within IoT-based smart environments. Our research centers on creating models resilient to attacks and robust against environmental uncertainties and dynamics, aiming to sustain high-quality monitoring and ensure energy sustainability within sensor networks.

\subsection{Decision-Theoretic or DRL-based Monitoring Systems} 

\citet{eichner2023optimal} proposed a method for optimally placing vibration sensors on offshore structures, aiming to maximize the monitoring system's data value. This method approaches sensor placement as a sequential decision-making issue, guided by a heuristic based on available data for inspections and repairs. \citet{nefedov2023model} developed an environmental monitoring system using multicriteria discrete optimization to fulfill key requirements like low power consumption and accuracy, employing a WSN guided by a generalized model for operation under uncertainty. Additionally, \citet{giordano2023value} introduced a decision theory framework to evaluate how prior knowledge of seismic activities affects the Value of Information (VoI) in Structural Health Monitoring Systems (S2HM), focusing on uncertainties and seismic hazards.

\citet{yun2022cooperative} introduced a multiagent deep reinforcement learning (DRL) strategy for industrial surveillance in smart cities, focusing on autonomous UAV communication management for consistent network coverage. Similarly, \citet{nguyen2023utilizing} applied DRL to improve environmental monitoring systems, enabling UAVs to collaborate, share data, and reposition for better connectivity. \citet{nguyen2021federated} developed DeepMonitor, a traffic monitoring system using Software-Defined Networking (SDN) and IoT edge nodes, employing the double deep Q-network (DDQN) for optimal flow rule policies. Furthermore, \citet{sultan2021energy} proposed a method using long short-term memory DQN for sensor selection in IoT tracking, incorporating the Kalman filter to mitigate measurement noise effects on distance estimation.  

Despite advancements in DT and DRL for monitoring system efficiency and performance, research on energy consumption balance and sustainable, resilient monitoring in resource-limited environments remains scant. Our work addresses not only cyberattack resilience in monitoring systems but also proposes adaptive solutions for dynamic smart environments.

\section{Problem Statement}  \label{sec:problem-statement}
\begin{figure}[t]
\centering
\includegraphics[width=0.8\textwidth]{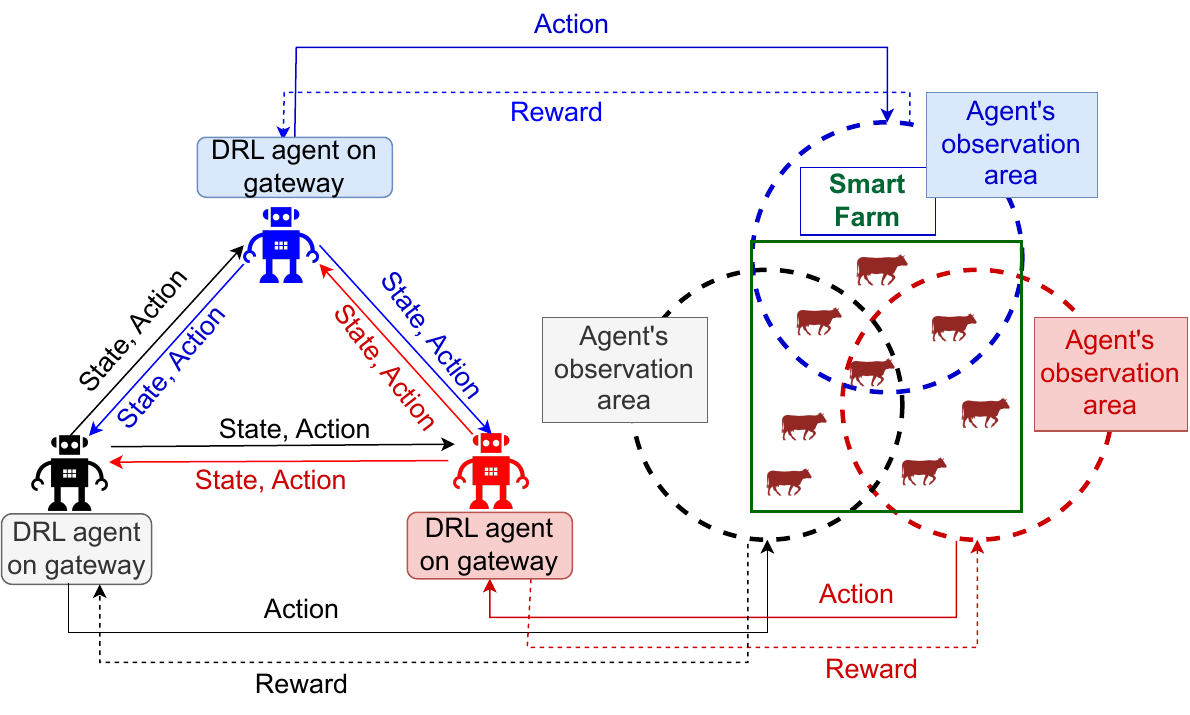}
\vspace{-4mm}
\caption{The considered multi-agent DRL environment.}
\label{fig:DRL}
\vspace{-4mm}
\end{figure}

To enhance animal condition monitoring and prolong system lifespan, the smart farm utilizes multi-agent DRL (depicted in Fig.~\ref{fig:DRL}). DRL agents operating on a LoRa gateway seek an optimal policy for selecting low-energy sensors to forward data to high-energy counterparts within each time interval. Specifically, these agents strive to optimize both monitoring quality and the residual energy of solar-powered sensors in the smart farm setting. The DRL agents aim to achieve the following:
\begin{eqnarray}
\label{eq:OF}
\argmax_{\mathbf{a^*}} \sum_t^T w_{MQ}\mathcal{MQ}(a_t) + w_{RE} \mathcal{RE}(a_t). 
\end{eqnarray}
Here, $\mathbf{a^*}$ represents the optimal action set $a_t$ chosen by an agent to maximize the sum of monitoring quality $\mathcal{MQ}(a_t)$ and average remaining energy level $\mathcal{RE}(a_t)$ of sensor nodes at time $t$, with both $\mathcal{MQ}(a_t)$ and $\mathcal{RE}(a_t)$ normalized to the range $[0,1]$ and equally weighted by $w_{MQ}=w_{RE}=0.5$. The formula for $\mathcal{MQ}(a_t)$ is:
\begin{eqnarray}
\label{eq:MQ-at}
\mathcal{MQ}(a_t) = \frac{\sum_{t=1}^{T_{current}(a_t)}\sum_{i=1}^X\sum_{j=1}^d \mathrm{mq} (i, j)}{X \times d},
\end{eqnarray}
where $T_{current}(a_t)$ denotes the operation time step under action $a_t$, $X$ represents the total number of sensed data points, and $d$ is the count of attributes per animal as described in the Node Model (see Section ~\ref{subsec:node_model}). $GT_{i,j}$ refers to the ground truth data for the $j$th attribute of the $i$th data point, while $x_{i,j}$ denotes the observed data for these parameters. The function $\mathrm{mq}(i, j)$ measures the monitoring quality for the $j$-th attribute against the $i$-th ground truth data, returning 1 for exact matches ($x_{i,j} == GT_{i,j}$) and 0 otherwise. While it is recognized that ground truth data is unavailable during real-world deployment due to practical constraints—since data collection occurs only at the gateways—we assume its availability during the training phase to enable the development of DRL agents in a controlled or simulated environment. This approach of integrating simulated and real-world data, a standard practice in DRL research, allows agents to learn optimal actions under ideal conditions and enhances their ability to generalize to real-world scenarios \cite{osinski2020simulation, kang2019generalization, cutler2015efficient}. $\mathcal{RE}(a_t)$ is defined as:
\begin{eqnarray}
\label{eq:re-at}
 \mathcal{RE}(a_t) = 1 - (\mathcal{E}_{SG} + \mathcal{E}_{SS} + \mathcal{E}_{active} +\mathcal{E}_{sleep}). 
\end{eqnarray}
The definition of $\mathcal{RE}(a_t)$ involves energy consumption of data transmission from sensors to sensors ($\mathcal{E}_{SS}$) and from sensors to gateways ($\mathcal{E}_{SG}$), sensors' energy drained in active ($\mathcal{E}_{active}$) and sleep ($\mathcal{E}_{sleep}$) mode.  We focus exclusively on the energy consumption of sensor communications for data transmission, as other communication tasks (e.g., sending requests) consume significantly less energy in comparison.  Notably, enhancing $\mathcal{MQ}(a_t)$ and $\mathcal{RE}(a_t)$ represents conflicting goals; for instance, activating more sensors to increase data coverage ($\mathcal{MQ}(a_t)$) will, in turn, reduce the overall energy efficiency ($\mathcal{RE}(a_t)$) due to the greater energy expenditure on data transmission. The smart farm system is designed with key components: effective energy policy learning, monitoring mechanisms (e.g., DRL-based monitoring using either transfer learning or decision theory or other heuristics; see Section~\ref{sec:exp-setup}), cyber-attack resilient data aggregation and updating, and a mechanism to identify deceptive data. These components are elaborated in Section~\ref{sec:proposed-scheme}.

\section{System Model} \label{sec:system-model}
\begin{figure}[t]
\centering
\includegraphics[width=0.6\textwidth]{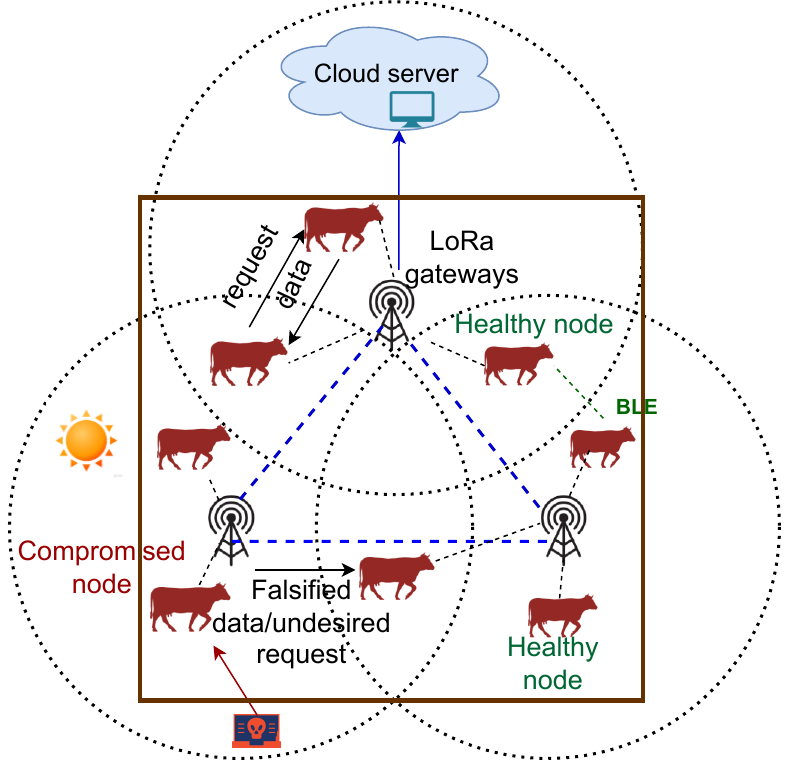}
\vspace{-3mm}
\caption{The considered wireless solar sensor-based smart farm network.}
\label{fig:NM}
\vspace{-3mm}
\end{figure}

\subsection{Network Model} \label{subsec:network_model}

The proposed smart farm system integrates a network comprising solar-powered sensors, LoRa gateways, and a cloud server, as depicted in Fig.~\ref{fig:NM}. These sensors periodically gather and transmit data on animal conditions either directly to gateways or to other sensors upon request. Gateways serve as intermediaries, aggregating data to reflect average animal conditions before relaying it to the cloud server at set intervals. LoRa technology is strategic, minimizing IoT connectivity costs and enhancing range.  In our implementation, we use three gateways as an example configuration that can fully cover the operating area of our farm to demonstrate the system's functionality. The appropriate number of gateways required for a farm, however, may vary depending on the farm's size, layout, and specific needs. For practical deployments, the number of gateways should be determined based on factors such as the farm's geographical area, the density of IoT devices, and the communication range of the gateways.

To attain autonomous monitoring capability, we leverage DRL algorithms operating on the gateways.  A multi-agent approach is used to increase the system's scalability as three agents fully cover the farm. By having multiple agents make decisions in parallel, multi-agent DRL helps in scenarios where a centralized decision-making process is inefficient or impractical.  To expedite the learning process in DRL, we incorporate transfer learning (TL) supported by the multi-agent approach and decision theory (DT) into DRL, which can significantly reduce the convergence time.  Given the resource constraints of the IoT environment, where data security measures like encryption are impractical due to limited computational and communication resources, the network is particularly vulnerable to cyberattacks. These potential security breaches, detailed in Section~\ref{subsec:attack-model}, could compromise the integrity of data transmission between sensors and between sensors and gateways. The design prioritizes optimizing energy use among the solar-powered sensors, which are the network's sole energy-limited components. Our protocol design balances monitoring efficacy with energy conservation, safeguarding against these vulnerabilities.

\subsection{Node Model} \label{subsec:node_model}

In the smart farm environment, sensors are designed to exchange data both on-demand and at scheduled intervals with LoRa gateways, considering the variable energy levels of solar-powered sensors influenced by animal movements and weather conditions. Each sensor node $i$ at time $t$, denoted as $\mathrm{sn}_t^i$, is characterized by four specific attributes, $
\mathrm{sn}_t^i = [\mathrm{temp}_t^i, \mathrm{hb}_t^i, \mathrm{ma}_t^i, \mathrm{bl}_t^i]$,
where $\mathrm{temp}_t^i$ represents the temperature, $\mathrm{hb}_t^i$ the heartbeat, $\mathrm{ma}_t^i$ the movement activity, and $\mathrm{bl}_t^i$ the battery level of sensor node $i$ at time $t$. The attributes $\mathrm{ma}_t^i$ and $\mathrm{bl}_t^i$ are normalized within the range $[0,1]$, indicating percentages. While gateways have ample storage for data collected during the monitoring period, sensor nodes face storage limitations. Data is timestamped, and when storage capacity is exceeded, the oldest data is deleted for new data. The system categorizes sensors as high-energy sensors (HESs) and low-energy sensors (LESs) based on their remaining energy levels.

HESs relay their collected data directly to LoRa gateways, whereas LESs communicate their data to nearby HESs using Bluetooth Low Energy (BLE) for efficiency. This smart farm employs the LoRa protocol for extended-range communications, covering distances from 5 to 15 $km$ at speeds up to 27 $kbps$, and BLE for shorter ranges, up to 100 meters, with faster data transfer rates of 2 $Mbps$. Energy usage is a critical consideration: the LoRa radio on SAM R34/35 models uses about 170 $mW$ for data transmission, in stark contrast to the BLE radio's 11 $mW$~\cite{manualR3, manualR2}. This discrepancy means that transmitting data via LoRa consumes approximately 1,100 times more energy than via BLE. The ratio is determined based on power dissipation during transmission and data rate for LoRa and BLE. The power dissipation values are \( P_{\text{LoRa}} = 170 \) mW for LoRa and \( P_{\text{BLE}} = 11 \) mW for BLE. The data rates are \( R_{\text{LoRa}} = 27 \) Kb/s (\( 27,000 \) b/s) for LoRa and \( R_{\text{BLE}} = 2 \) Mb/s (\( 2,000,000 \) b/s) for BLE.  The energy required to transmit a single bit is given by:
\begin{equation}
    E = \frac{\text{Power}}{\text{Data Rate}}
\end{equation}

Converting to standard units:

\begin{equation}
    E_{\text{LoRa}} = \frac{170 \times 10^{-3} \, \text{W}}{27,000 \, \text{b/s}} \approx 6.296 \times 10^{-6} \, \text{J/bit}
\end{equation}

\begin{equation}
    E_{\text{BLE}} = \frac{11 \times 10^{-3} \, \text{W}}{2,000,000 \, \text{b/s}} = 5.5 \times 10^{-9} \, \text{J/bit}
\end{equation}

The ratio of energy consumption per bit between LoRa and BLE is:

\begin{equation}
    \text{Ratio} = \frac{E_{\text{LoRa}}}{E_{\text{BLE}}} = \frac{6.296 \times 10^{-6}}{5.5 \times 10^{-9}} \approx 1145
\end{equation}

Initially, each sensor node starts with a 5 $kW$ power reserve. Solar charging rates vary with light exposure, yielding about 10 $mW/cm^2$ outdoors and 0.1 $mW/cm^2$ indoors~\cite{kim2014ambient, mathuna2008energy}.

\subsection{Attack Model} \label{subsec:attack-model}

Many works discuss detailed insights into the conditions under which cyber attacks can be executed, even in scenarios with local control and ownership. They outline that attackers with advanced capabilities or prolonged access to physical locations could exploit vulnerabilities, demonstrating the importance of robust security measures even in seemingly secure environments~\cite{gupta20, sontowski2020, yazdinejad2021review}. Thus, our work considers the following attack behaviors in the smart farm system: 
\begin{enumerate}
\item {\bf False data injection attacks}: Compromised sensors may transmit altered data to gateways or other sensors. In man-in-the-middle attacks, unauthorized entities can intercept and manipulate data sent to HES nodes.  For instance, attackers could modify an animal's body temperature or heart rate readings to skew monitoring results, leading to incorrect actions by the system. These attacks compromise the accuracy and reliability of the collected data and may disrupt decision-making processes in smart farming applications.
\item {\bf Non-compliance to protocol}: A compromised sensor may deviate from the data transmission protocol. For instance, an LES could ignore data requests or withhold transmissions to conserve energy. Similarly, a compromised HES might deny requests from sensor node $A$ or transmit data from sensor node $B$ instead.  This non-compliance undermines the integrity and consistency of data flow, potentially creating gaps in monitoring or incorrect analyses.
\item {\bf Denial-of-service (DoS)}: DoS attacks flood HES nodes with false requests, leading to unnecessary data transmissions and rapid energy depletion. In critical systems like smart farm monitoring, this could lead to system downtime, degraded service quality, and delayed responses to real-time events.
\item {\bf Sensor data obstruction}: Deauthentication attacks block sensors from connecting to the network, disrupting communication with other sensors and LoRa gateways~\cite{wang2020slora}.  Such disruptions create blind spots in the monitoring system, as vital data from disconnected sensors is no longer available, affecting overall system accuracy and responsiveness.
\item {\bf Neural trojan attack}: A compromised gateway may use a Trojan trigger to alter the neural network (NN) model, leading to suboptimal policies generated by the DRL agent~\cite{liu18-Purdue}.
\item {\bf Fast Gradient Sign Method (FGSM)}: This attack generates adversarial inputs to maximize the loss function, misleading DRL agents into taking unfavorable actions~\cite{fel14}.
\item {\bf Projected Gradient Descent Attack (PGDA)}: Using the PGD optimization technique, this attack misleads DRL agents by inducing misclassification through iterative gradient adjustments, degrading the monitoring quality and system performance~\cite{ayas22}.
\end{enumerate}

These attacks demonstrated in real-world smart farms, and IoT contexts underscore the susceptibility of sensor nodes and LoRa gateways~\cite{kuntke2022lorawan}. Attack severity is assessed based on the probability of each attack type occurring.  During evaluation, this probability represents the chance of an attack being successfully executed, potentially disrupting the operation of the smart farm system.

Our work seeks to enhance smart farm system resilience against cyber and adversarial attacks through trustworthy evidence aggregation and the proposed DT-guided DRL approach rather than relying on conventional defense mechanisms. This ensures that our system is fault-tolerant, maintaining normal operations even in the face of threats.  Our proposed uncertainty-aware aggregation method (see Section~\ref{subsec:deceptive-detect}) is designed to filter out falsified or modified data injected by attackers. The DRL-based monitoring system enhances robustness when facing abnormal behaviors triggered by attacks 2--4. Moreover, the DT-guided DRL enables agents to effectively mitigate attacks targeting the DRL process (i.e., attacks 5-7). This is because the DT component of the agents operates without a learning process, making it less susceptible to adversarial attacks than fully DRL-based approaches.

\subsection{Energy Model} \label{subsec:energy-policy}
As detailed in Section~\ref{subsec:node_model}, sensor nodes are classified as high-energy sensors (HESs) and low-energy sensors (LESs) according to their energy levels, which dictates their capacity to transmit data directly to LoRa gateways. We use $L_{bl}$ to denote the minimum energy threshold required for a sensor to employ the LoRa protocol for communication. A sensor with an energy level above $L_{bl}$ is identified as a HES and periodically sends data to the gateways (e.g., every 30 seconds). In contrast, sensors with energy below this threshold are designated as LESs and route their data through a nearby HES upon request. We introduce a sophisticated energy policy for data transmission to enhance the sensors' energy efficiency and extend the overall system lifetime, detailed in Section~\ref{sec:proposed-scheme}. 


In our smart farm system, the energy level of a sensor node is normalized to the range $[0, 1]$, where 1 represents a fully charged node. We categorize energy usage into four types associated with data transmission and natural depletion over time.  HESs, with energy levels above a defined threshold $L_{bl}$ (e.g., 0.3), transmit data to gateways using the LoRa protocol at intervals $T_u$ (e.g., 30 sec.). Conversely, LESs with energy below $L_{bl}$ send their data to a nearby HES via BLE.


\section{DT-guided DRL-based Monitoring}\label{sec:proposed-scheme}

In our smart farm model, environmental changes notably impact monitoring efficiency and energy usage. Solar-powered sensor node charging efficiency decreases during overcast weather or when animals move to sheltered areas, affecting energy availability and monitoring quality.  For instance, on a sunny day, the solar panels on the sensor nodes charge efficiently, providing ample energy for continuous monitoring of the animals' movement, temperature, and feeding patterns. However, when the animals move to sheltered areas to avoid rain, the solar panels attached to these shelters may not receive direct sunlight, further decreasing charging efficiency. This compounded effect can reduce the system's ability to provide accurate monitoring and potentially delay the detection of any abnormal behaviors or health issues in the animals.  Traditional fixed or rule-based energy management approaches are inadequate for these dynamic conditions. To address this, we devised a dynamic energy management strategy for sensor data transmission. This strategy incorporates a DRL agent to optimize monitoring quality while maintaining sensors' sufficient energy levels.

The motivation for DT-guided DRL stems from combining the strengths of DT and DRL. DT provides heuristic decision-making based on environmental awareness, albeit with limited complexity. In contrast, DRL offers greater flexibility with DNNs but requires extensive data and interactions for pattern recognition. DT-guided DRL integrates these algorithms in a relay-like fashion, mitigating cold start issues and optimizing performance through their complementary advantages. Furthermore, we introduce an uncertainty-aware method for monitoring opinion updates, capable of identifying deceptive data from adversarial attacks, thereby enhancing system resilience.

\subsection{DRL-based Monitoring} \label{subsec:drl-based-monitoring-system} 

We implemented a dynamic energy policy tailored for sensor data transmission. This policy is optimized by a DRL-based agent, enabling the fine-tuning of both energy efficiency and monitoring quality. Such optimization allows for immediate adjustments in response to environmental variations, ensuring the system remains responsive and maintains performance amidst fluctuating conditions. 

\subsubsection{\bf State space $(\mathcal{S}_t)$}  We maintain a local database to store all received data from sensor nodes on each LoRa gateway. The DRL agents will compute and update the corresponding opinions about the sensed data of animal conditions. Each DRL agent deployed on the gateway can only obtain its own database and has limited monitoring ability to observe the environment. A DRL agent can also observe the remaining energy level of each sensor node located in its transmission range. We define the state space for an agent $i$ by $\mathcal{S}_i = \{s_i^1, s_i^2, \ldots, s_i^t\}$, where $t$ is the current time step of the smart farm operation.  Each state in state space $\mathcal{S}_i$ is denoted by $s_i^t = \{\{\mathrm{re}_1^t, \mathrm{o}_1^t\}_i, \ldots, \{\mathrm{re}_j^t, \mathrm{o}_j^t\}_i\}$, where $\mathrm{o}_j^t$ is the opinion computed for animal $j$ in agent $i$'s transmission range at time $t$, and $\mathrm{re}_j^t$ is a list of remaining energy on each sensor close to a sensor (i.e., animal) $j$ at time $t$. 

\subsubsection{\bf Action space $(\mathcal{A}_t)$} The objectives of our proposed smart farm monitoring system are to maximize the monitoring quality of animals’ conditions and the lifetime of the system. In the WSN-based system, the lifetime of the whole system significantly depends on the energy level of sensor nodes. To achieve both objectives, we determine a percentage threshold $\rho$ and compute the list of LESs ranked by their remaining energy level in ascending order. Only the top $\rho$ percentage of sensor nodes from the list can transmit their data to the nearby HES in each time interval. Once the initial value of $\rho$ is provided, the DRL agents will dynamically adjust it to learn an optimal value of $\rho$.  We define the action space with three discrete actions at time $t$, $\mathcal{A}_t = \{\mathrm{increase}, \mathrm{decrease}, \mathrm{stay}\}$ with the increment or decrement with $\tau$ (e.g., 0.05 where the $\rho$ is scaled in $[0, 1]$ as a real number), in which the DRL agent can select an action in each step per update interval. DRL agents must identify the optimal $\rho$ because more sensor nodes can send their data to the gateways with a larger $\rho$, resulting in high monitoring quality of animal status at the cost of depleting the remaining energy level, and vice versa. We use a single threshold $\rho$ instead of a threshold for each sensor because controlling the number of sensor nodes transmitting their sensed data in a time interval is more efficient. In addition, it contributes to guaranteeing the sensors will change by taking the corresponding actions.

\subsubsection{\bf Reward} The DRL agent will receive the immediate reward after taking action at time $t$, formulated by $r_t = \mathcal{MQ}(a_t) + \mathcal{RE}(a_t)$ (see Eq.~\eqref{eq:MQ-at} and Eq.~\eqref{eq:re-at}). The DRL agent will select an action to maximize the accumulated expected return, formulated by $\mathcal{R}_t = \sum_{t=0}^T \gamma^t r_t$, where $r_t$ is a reward at time $t$, $\gamma$ is a discount factor (i.e., 0.9), and $T$ is the period of an episode.

Our proposed DRL-based approach inherently addresses scalability challenges by decoupling the computational cost from the number of nodes in the system. The computational cost in DRL primarily arises from the size of the action space, i.e., the number of possible actions the agent can take. In typical DRL systems, as the number of nodes (or animals in a farm) increases, the action space also grows, which can lead to higher computational demands. However, our design significantly reduces this challenge by limiting the action space to only three possible actions, simplifying the decision-making process. This combination of reduced action space and a DT-guided approach makes our approach particularly suitable for large-scale deployment in rural agricultural settings, where computational and energy resources are often limited. Moreover, this DRL-based approach is designed to be energy-adaptive, meaning that the system can select the optimal action based on the current environmental conditions, including data and energy loss due to weather changes or sensor malfunctions. Specifically, the approach dynamically adjusts its decisions to optimize energy consumption and maintain functionality despite environmental disturbances, ensuring that the system remains efficient and resilient even when the availability of sensor data and energy fluctuates. 

\subsection{Integration of DT with DRL}

\begin{figure*}[t]
\centering
\includegraphics[width=0.9\textwidth, height=0.6\textwidth]{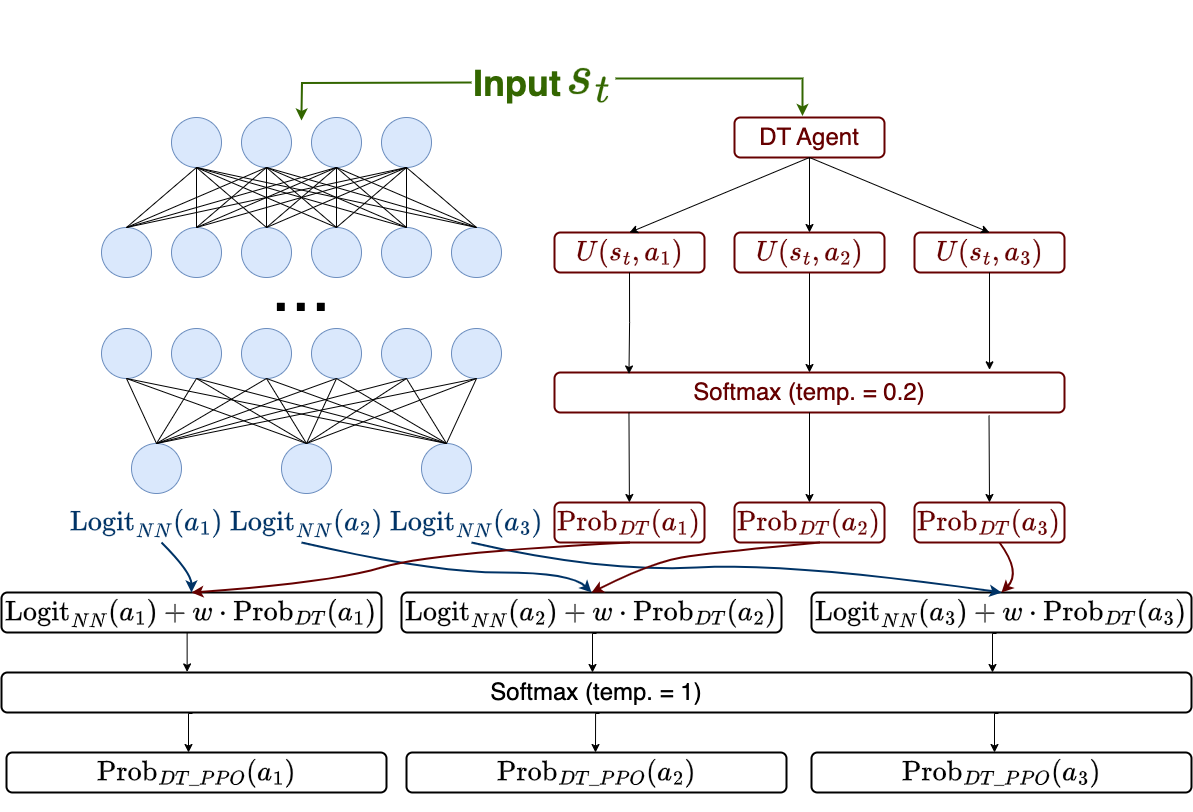}
\vspace{-4mm}
\caption{Decision-making procedure of a DT-guided DRL agent.}
\label{fig:dt-drl}
\vspace{-4mm}
\end{figure*}

\subsubsection{\bf DT-based Monitoring}
In our system, agents on each gateway strive to identify the most effective energy policy that enhances monitoring quality while ensuring adequate remaining energy under cyber threats and energy consumption challenges, as outlined in Eq.~\eqref{eq:OF}. To accomplish this, we introduce a utility function that guides each agent to choose actions maximizing this utility, represented by:
\begin{equation}
\label{eq:dt-utility}
U(s_t, a_t) = w_1 \cdot A(s_t, a_t) + w_2 \cdot E(s_t, a_t), 
\end{equation}
where $A(s_t, a_t)$ represents the action's impact on monitoring quality and $E(s_t, a_t)$ on the remaining energy, given the current state $s_t$ and action $a_t$. These impacts are quantified by combining the average values of monitoring quality and remaining energy from previous states with the change in a specific factor $\rho$ after executing the action. Both metrics, $A(s_t, a_t)$ and $E(s_t, a_t)$, are normalized within the $[0, 1]$ range. Weights $w_1$ and $w_2$ are adjusted to equally value monitoring quality and energy preservation, fine-tuned through empirical analysis where $w_1+w_2=1$. Based on this utility function, DT agents will opt for the action from the set $\mathcal{A}$ that yields the highest utility, ensuring a balance between system performance and energy efficiency.  While DT agents use a utility function akin to a reward function, they differ from DRL in that DT typically assumes the agent has prior knowledge of the environment and its outcomes. This assumption makes DT more static, as it does not account for dynamic or uncertain environmental changes. In contrast, DRL relies on exploration, assuming the agent has limited or no prior knowledge of the environment. As a result, DT agents may outperform DRL-based agents at the initial stages of system operation, where DRL agents begin with random actions.

\subsubsection{\bf Integrating DT with DRL}

The DT-guided DRL's effectiveness lies in combining the DT agent's action probabilities with the neural network (NN)'s output (see Fig.~\ref{fig:dt-drl}). The process unfolds in several steps. \paragraph{\bf Generation of action distributions} Each DT agent computes the utility of possible actions in $\mathcal{A}$ (see the utility function in Eq.~\eqref{eq:dt-utility}). To integrate the DT into a DRL agent's decision-making, we must first convert the outcome of the utility function into a distribution, similar to how DRL agents select actions based on an action distribution. This ensures consistency in decision-making across both systems.  A \textit{softmax} function then transforms these utility values into a probability distribution $\mathrm{Prob}_{DT}(a^*)$, where actions with higher utilities are more likely to be chosen. Adjusting the \textit{softmax} temperature enhances the DT agent's decisiveness. The DRL distribution by a DNN reflects current environmental learning, with the Proximal Policy Optimization (PPO)~\cite{schu17} selected for its outperformance. \paragraph{\bf Integrating DT and PPO agents}  This merges their \textit{logits} and probability distribution. The DRL agent's $\mathrm{Logit}_{NN}(a^*)$ is obtained directly from its output, whereas the DT agent's $\mathrm{Prob}_{DT}(a^*)$ is computed from the previous step. The two distributions are then summed, applying a weight $w$ to $\mathrm{Prob}_{DT}(a^*)$ to adjust the DT agent's influence, resulting in $\mathrm{Logit}_{NN}(a^*)+w\cdot\mathrm{Prob}_{DT}(a^*)$.  Initially, $w$ is set to 1, and it decreases over time (i.e., decay rate with 0.0003), allowing the DNN's increasingly accurate policy to gradually take precedence in decision-making. 
\paragraph{\bf Softmax layer addition} To ensure the final action selection probabilities are valid, a \textit{softmax} function with a temperature setting of 1 is applied to the integrated outputs, a standard practice in NN training. 

\paragraph{\bf Training the DT-guided DRL agent} With complete integration, the agent undergoes training using traditional DRL methods like DQN and PPO, learning the optimal policy by interacting with the environment in real-time.

This methodology capitalizes on the strengths of both decision theories and deep learning, ensuring a more reliable and effective learning process from the outset.

\subsection{Aggregation of Uncertain Observations} \label{subsec:uncertainty-aware-monitoring-system}
Considering the potential cyber-attacks in the monitoring system that could introduce uncertainty, we propose an uncertainty-aware deceptive data detection method that quantifies this uncertainty using the concept of {\em Subjective Logic} (SL)~\cite{sj16}.  After receiving sensed data from sensor nodes consisting of attributes described in Section~\ref{subsec:node_model}, the DRL agents will aggregate the gathered observations with an uncertainty-aware technique stemming from a lack of evidence. The animal conditions can be clustered into $K=3$ classes based on the range of each attribute, whose average value we can estimate along with the likelihood of each class as well as their uncertainty levels. For example, one piece of sensed data about animal temperature can be recorded as one of the ranges: 37 or below as lower than normal, 38 to 40 as normal, and 42 or above as higher than normal. 

The DRL agents will formulate an opinion about animal conditions by leveraging a belief model from SL to explicitly address different uncertainty types. In SL, an agent $A$ computes an opinion for a given proposition $X$, denoted by $\omega^A_X = \{\bm{b}_X, u_X, \bm{a}_X\}$, where $\bm{b}_X$ is belief masses distribution, $u_X$ is the uncertainty mass and $\bm{a}_X$ is the base rate (i.e., prior belief) distribution of a random variable $X$ in domain $\mathbb{X}$. The components satisfy the additivity requirement with $u_X + \sum \bm{b}_X(x) = 1$.  Without prior knowledge, we initialize the base rate equally for each belief mass, i.e., $a_X (x_i) = 1/K$ for any $x_i$ (i.e., uniform distribution).  Each instance of sensed data will be regarded as evidence to update the opinions. This work considers uncertainty introduced by a lack of evidence, called {\em vacuity} in SL. Updating opinions in SL will stop when uncertainty reaches zero, hindering further updates with new evidence. To avoid this, we will deploy the uncertainty (vacuity) maximization (UM) technique to reduce the effect of conflicting evidence by transforming it into the vacuity of an opinion.  The UM computation~\cite{sj16} is given by:
\begin{eqnarray}
\ddot{u}_X = \min_i \left[ \frac{{\bf P}_X(x_i)}{{\bf a}_X (x_i)} \right], \; \text{for}\; x_i \in \mathbb{X},    
\end{eqnarray}
where $\ddot{u}_X$ is uncertainty-maximised $u_X$ and ${\bf P}_X(x_i)$ is the projected probability distribution, given by ${\bf P}_X(x_i) = {\bf b}_X(x_i) + {\bf a}_X(x_i)u_X$. 


\subsection{Detection of Deceptive Data} \label{subsec:deceptive-detect}

To counteract the impact of poisoned data from adversarial attacks, such as false data injection, on monitoring quality, our system incorporates the concept of {\em degree of conflict} (DC) from {\em Subjective Logic} (SL)~\cite{sj16}. This approach aids in identifying deceptive data as gateways collect information from potentially compromised sensors. We employ a measure of {\em projected distance} (PD) to evaluate discrepancies between differing opinions, defined as follows:
\begin{eqnarray}
\widehat{PD}(\omega^A_X) &=& \frac{\sum_{j \in B} PD(\omega^A_X,\omega^j_X)}{|B|}, \\
\text{where} \; PD(\omega^A_X,\omega^j_X) &=& \frac{\sum_{x\in \mathbb{X}} |\omega^A_X(x) - \omega^j_X(x)|}{2}, \nonumber    
\end{eqnarray}
where $\omega^A_X$ represents the opinion of agent $A$ and $\omega^j_X$ the opinions of other agents $j \in B$, regarding the proposition $X$. This calculation occurs after a gateway updates its opinion based on newly received data. Should the resulting PD surpass a predefined threshold $\phi \in [0,1]$, the data from agent $A$ is deemed deceptive and consequently disregarded. This method provides a systematic way to filter out unreliable data, enhancing the integrity and accuracy of the monitoring system.

 	\begin{table}[t]
\centering
\caption{\sc EVD Dataset Description}
\label{tab:evd_data}
\vspace{-2mm}
\begin{tabular}{|P{4cm}|P{8cm}|}
        \hline
        \textbf{Metric} & \textbf{Description} \\
        \hline
        Serial & A unique animal identifier \\
        \hline
        HR & Heart Rate of the animal \\
        \hline 
        Average temperature & Average body temperature in Celsius \\
        \hline
        Min-temperature & Minimum temperature in Celsius \\
        \hline
        Max-temperature & Maximum temperature in Celsius \\
        \hline
        Average-activity & Average activity recorded by the number of steps taken \\
        \hline
        Battery-level & Residual battery life \\
        \hline 
        Timestamp & Date and time of transmission \\
        \hline 
    \end{tabular}
    \vspace{-4mm}
\end{table}

\section{Simulation Setup} \label{sec:exp-setup}
\subsection{Parameterization} \label{subsec:param}

In this study, we utilize the SmartFarm Innovation Network$^{\mathrm{TM}}$ at Virginia Tech as a primary source for collecting and analyzing agricultural data across Virginia. This platform facilitates access to datasets from a smart farm managed by the College of Agriculture and Life Sciences at Virginia Tech. The data are collected from various devices, such as EmbediVet Implantable Temperature Devices (EVD), Halter Sensors, Heart Rate Sensors, and Implantable Temperature Sensors, whose attributes are summarized in Table~\ref{tab:evd_data}. 

Animal movements are tracked using GPS receivers attached to wearable sensors on the animals, which communicate location data to microcontrollers and subsequently to gateways via LoRa~\cite{dos2021}. Compromised datasets are artificially generated to align with ground truth data, incorporating the attack behaviors outlined in Section~\ref{subsec:attack-model}. The compromised data is generated based on the real distributions of each animal attribute (e.g., body temperature), ensuring that all data points are entirely synthetic. While these data points appear plausible, they do not correspond to actual real-world observations.

The research setting is a 40-acre farm with a perimeter of 400 meters on each side, where 20 cows are monitored by three gateways to ensure comprehensive coverage. This simulation spans 24 hours, during which each gateway, equipped with a DRL agent, tracks various health indicators of each animal, including heart rate, temperature, and movement. The defined normal ranges for these indicators are set as $[37.8, 39.2]$ Celsius for temperature, $[48, 84]$ beats per minute for heart rate, and $[1, 2]$ meters per second for movement. Cow movement probabilities and speeds follow a normal distribution, with mean speeds at 1.5 m/s and a standard deviation of 0.1 m/s. The DRL agents are tasked with determining the optimal data transmission policy at 60-second intervals. The energy levels for the 5 HESs are fully charged (1.0), whereas the 15 LESs start with random energy levels between $[0.1, 0.2)$.  This study primarily focuses on ensuring normal system operations despite cyber threats rather than preventing intrusions into the smart farm system. We assume that 30\% of the sensors are initially compromised, and each gateway has a probability ($P_{AE}$) of being compromised. Parameters related to energy measurement are detailed in Section~\ref{subsec:node_model}.

Although real-world validation is not feasible at this stage, the datasets and hardware configurations used in this study are sourced from an actual farm, ensuring practical relevance. The SmartFarm Innovation Network at Virginia Tech serves as the real testbed, where only data collection has been conducted. Implementation within the testbed will be explored in future work.


\subsection{Metrics} \label{subsec:metrics}

We evaluate our simulation using the following metrics:
\begin{itemize}
    \item {\bf Accumulated Reward ($\mathcal{R}$):} This metric measures the total reward accumulated over all simulation runs, providing an overall assessment of the performance and efficiency of the deployed strategies, as outlined in Section~\ref{subsec:drl-based-monitoring-system}.
    \item {\bf Remaining Energy ($\mathcal{RE}$):} This metric evaluates the average remaining energy of the low-energy sensors (LESs) at each time interval, $T_u$, as defined by Eq.~\eqref{eq:re-at}, highlighting the energy efficiency of the monitoring system.
    \item {\bf Run Time ($R_T$):} This metric calculates the average training time per episode over 50 episodes, representing the computational efficiency and scalability of the learning algorithms.
    \item {\bf Monitoring Quality ($\mathcal{MQ}$):} Defined in Eq.~\eqref{eq:MQ-at}, this metric gauges the system's effectiveness in accurately monitoring and reporting on animal conditions, reflecting the overall monitoring accuracy.
\end{itemize}
These metrics collectively provide a comprehensive evaluation of the system's performance, encompassing its efficiency, energy management, operational speed, and data accuracy.

\subsection{Comparing Schemes} \label{subsec:comparing-schemes}

We evaluate the performance of the following schemes:
\begin{itemize}
    \item {\bf Proximal Policy Optimization (PPO)}~\cite{schu17}: DRL agents select optimal actions based on a learned policy. PPO uses an actor-critic style algorithm deploying multiple epochs of stochastic gradient ascent to update the policy. PPO is considered the baseline DRL since it performs best compared to other DRL schemes (e.g., DQN).
    \item {\bf TL-PPO-FT}: PPO-based DRL agents deploy a transfer learning (TL) algorithm that {\em fully} transfers knowledge from a pre-trained model with sufficient knowledge. The pre-trained model is trained in an environment resembling the multi-agent setup used for simulation.
    \item {\bf TL-PPO-PT}: PPO-based DRL agents learn model parameters from a pre-trained model that {\em partially} transfers knowledge to adapt to the environment under resource constraints. The partially pre-trained model is trained using a single-agent farm, which incurs lower computational costs than a multi-agent system. 
   \item {\bf DT-PPO}: DRL agents are trained using our proposed DT-guided DRL algorithm, which incorporates decision-theoretic principles to guide early exploration and improve sample efficiency while learning the policy.
   \item {\bf DT}: Agents select actions based on the highest utility value computed by our designed utility function in Eq.~\eqref{eq:dt-utility}, ensuring decision-making aligns with predefined task objectives.
    \item {\bf Adaptive-Energy-Distance (AED)}~\cite{alemayehu17}: Agents select actions to achieve both a high remaining energy level $(\mathcal{RE})$ and low total transmission distance $(\mathcal{D})$, thereby reducing data acquisition latency efficiently. The $(\mathcal{D})$ is defined as the total distance between each LES and its nearby HES, formulated by $\mathcal{D} = \sum_i d(p_i, q)$, where $p_i$ is the position of LES $i$, and $q$ is the position of LES $i$'s nearby HES. The agents select an action with the maximum $\mathcal{RE}-\mathcal{D}$.
    \item {\bf Random}: Agents randomly select actions from the action space at each step.
    \item {\bf Fixed-Energy (FE)}: Agents choose the top $30\%$ of sensor nodes with the least remaining energy.
\end{itemize}

As discussed in Section~\ref{sec:related-work}, we designed a rule-based heuristic, AED, based on the idea from~\cite{alemayehu17} as the only state-of-the-art approach against our proposed schemes.  DT-PPO is our proposed DT-guided DRL scheme, while others are baseline schemes for comparative performance analysis.  We evaluate the performance of the proposed DT-guided DRL schemes and the seven baseline schemes with 100 simulation runs based on the corresponding parameters and settings described in Section ~\ref{subsec:param}. For PPO, TL-PPO-FT, TL-PPO-PT, and DT-PPO, we use a batch size of 500 and a learning rate of 0.0008. We identify each hyperparameter by obtaining the optimal performance for each scheme.

\begin{figure*}[!ht]
  \centering
  \subfigure{
    \includegraphics[width=\textwidth, height=0.035\textwidth]{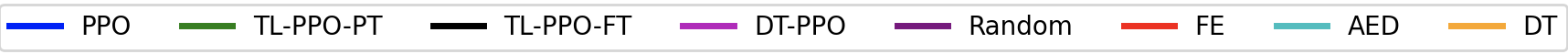}}
   \setcounter{subfigure}{0}
    \vspace{-3mm}
    
  \subfigure[\footnotesize{Accumulated Reward ($\mathcal{R}$)}]{
    \includegraphics[width=0.43\textwidth, height=0.3\textwidth]{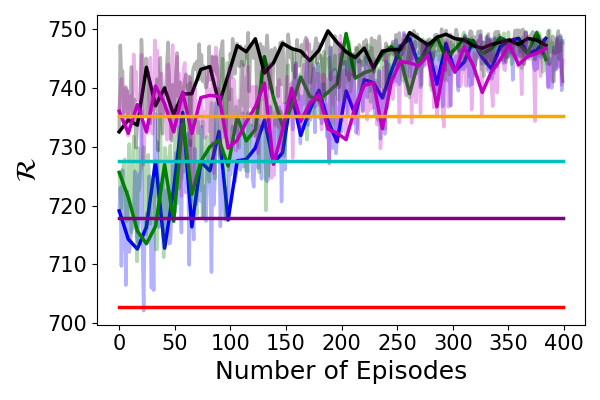}}
  \subfigure[\footnotesize{Monitoring Quality ($\mathcal{MQ}$)}]{
    \includegraphics[width=0.43\textwidth, height=0.3\textwidth]{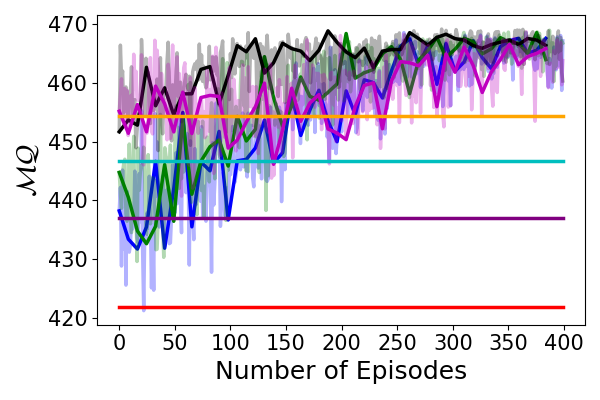}}
  \subfigure[\footnotesize{Remaining Energy ($\mathcal{RE}$)}]{
    \includegraphics[width=0.43\textwidth, height=0.3\textwidth]{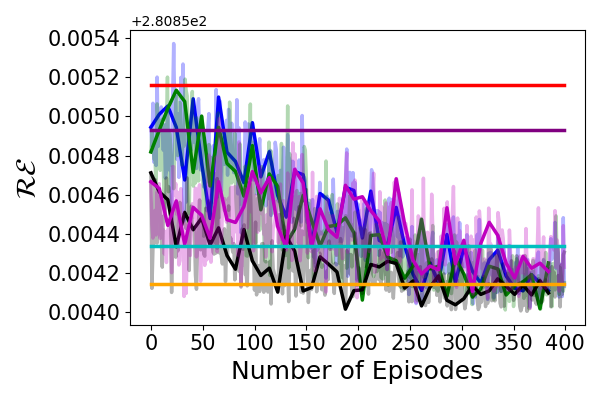}}
    \subfigure[\footnotesize{Run Time ($R_T$)}]{
    \includegraphics[width=0.43\textwidth, height=0.3\textwidth]{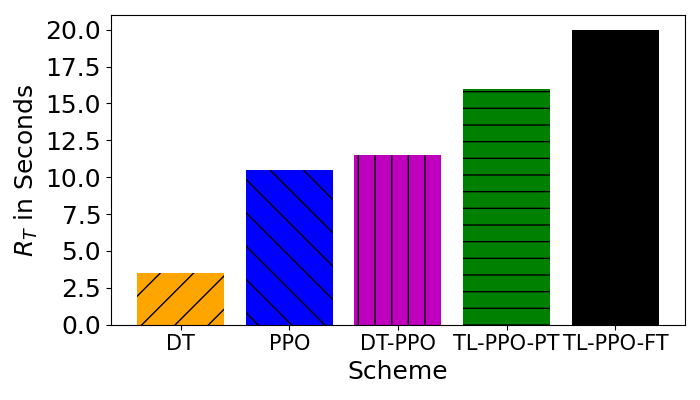}}
    \caption{Comparative performance analysis during training time with $P_A = 0.1$.}
\label{fig:cp}
\vspace{-4mm}
\end{figure*}

\section{Results and Analysis} \label{sec:results}

\subsection{Performance Comparison Analyses}

Fig.~\ref{fig:cp} showcases the learning progression across eight strategies, considering an attack probability ($P_A$) of 0.1. Strategies such as Random, FE (Fixed Energy), AED (Adaptive Energy Distance), and DT (Decision Theory) do not engage in a learning process, thus their performance across simulations is represented by constant horizontal lines, indicating no convergence time is applicable.

Our findings highlight that the DT-PPO (Decision Theory-guided - Proximal Policy Optimization) approach outperforms others like TL-PPO-PT (Transfer Learning - Partial Transfer) and standalone PPO in terms of accumulated rewards ($\mathcal{R}$) as seen in Fig.~\ref{fig:cp}(a), monitoring quality ($\mathcal{MQ}$) as in Fig.~\ref{fig:cp}(b), and run time ($R_T$) shown in Fig.~\ref{fig:cp}(d). This superior performance, however, inversely correlates with the remaining energy ($\mathcal{RE}$) metric in Fig.~\ref{fig:cp}(c), underscoring the inherent trade-off between striving for high monitoring quality and conserving energy within the system. While the runtime for TL-based methods includes both the training duration for the pre-existing model and the time required to adapt this model to new conditions, our findings indicate that the DT-PPO combination exhibits a significantly lower runtime than TL-based approaches. This efficiency stems from DT's ability to bypass the learning process and the need for additional computational efforts to gather feedback (such as rewards) from the environment.



Crucially, the DT-PPO scheme's comparative advantage shines in scenarios lacking pre-trained models suitable for transfer learning, directly addressing the limitation of transfer learning where the availability of a quality pre-trained model can be a significant barrier. Therefore, DT-PPO offers comparable performance without necessitating pre-existing models, making it particularly valuable in environments where such models are scarce or non-existent.

\subsection{Sensitivity Analyses}

\begin{figure*}[!ht]
  \centering
  \subfigure{
    \includegraphics[width=\textwidth, height=0.035\textwidth]{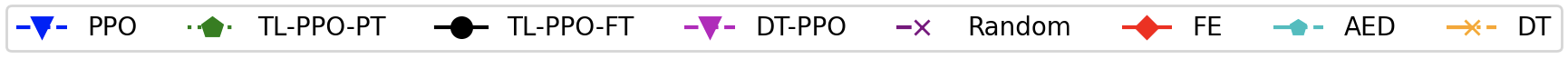}}
    \setcounter{subfigure}{0}    
  \subfigure[\footnotesize{Accumulated Reward ($\mathcal{R}$)}]{
    \includegraphics[width=0.43\textwidth, height=0.3\textwidth]{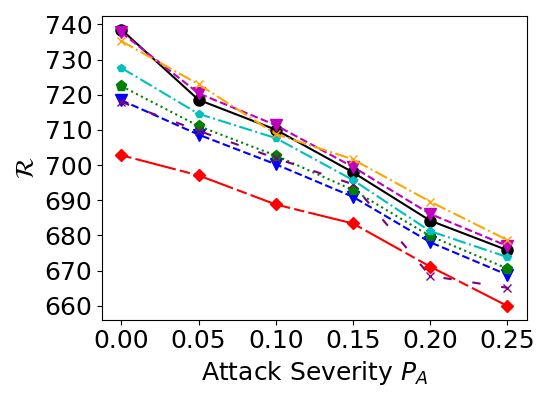}}
  \subfigure[\footnotesize{Monitoring Quality ($\mathcal{MQ}$)}]{
    \includegraphics[width=0.43\textwidth, height=0.3\textwidth]{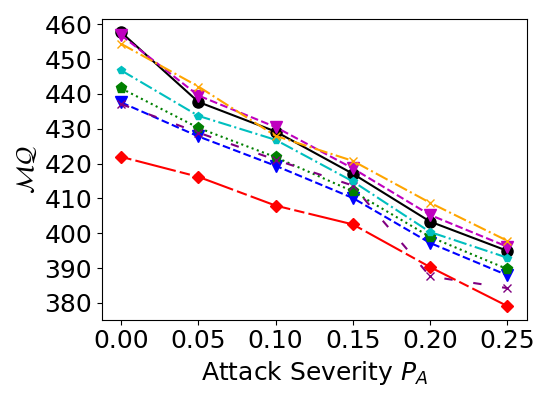}}
  \subfigure[\footnotesize{Remaining Energy ($\mathcal{RE}$)}]{
    \includegraphics[width=0.43\textwidth, height=0.3\textwidth]{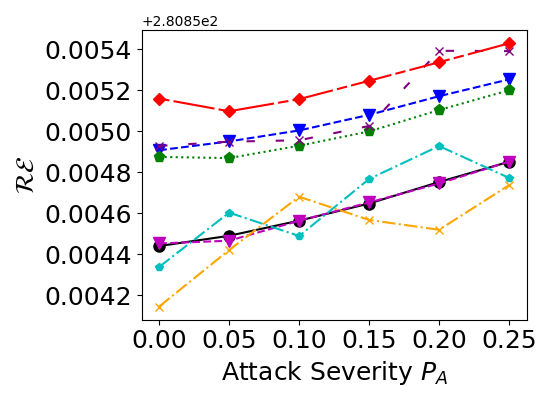}}
    \subfigure[\footnotesize{Run Time ($R_T$)}]{
    \includegraphics[width=0.43\textwidth, height=0.3\textwidth]{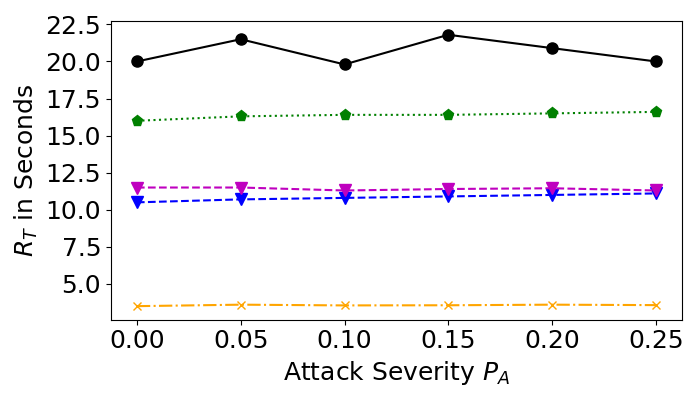}}
    
    \caption{Effect of Varying Attack Severity ($P_A$) on Solar Sensors.}
\label{fig:da}
\end{figure*}

\begin{figure*}[!ht]
  \centering
  \subfigure{
    \includegraphics[width=0.7\textwidth, height=0.035\textwidth]{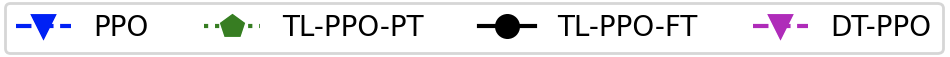}}
    \setcounter{subfigure}{0}
    \vspace{-2mm}
    
  \subfigure[\footnotesize{Accumulated Reward ($\mathcal{R}$)}]{
    \includegraphics[width=0.43\textwidth, height=0.3\textwidth]{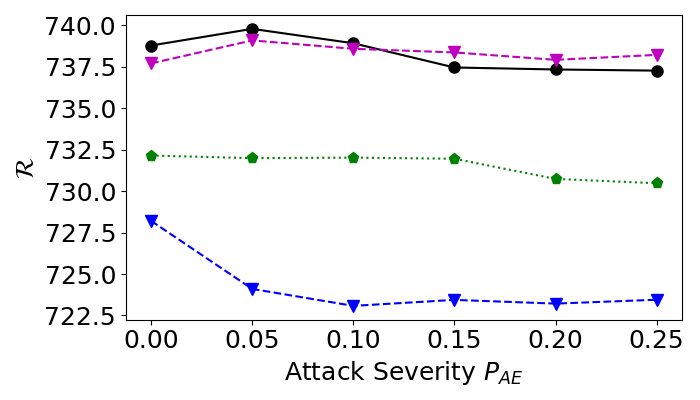}}
  \subfigure[\footnotesize{Monitoring Quality ($\mathcal{MQ}$)}]{
    \includegraphics[width=0.43\textwidth, height=0.3\textwidth]{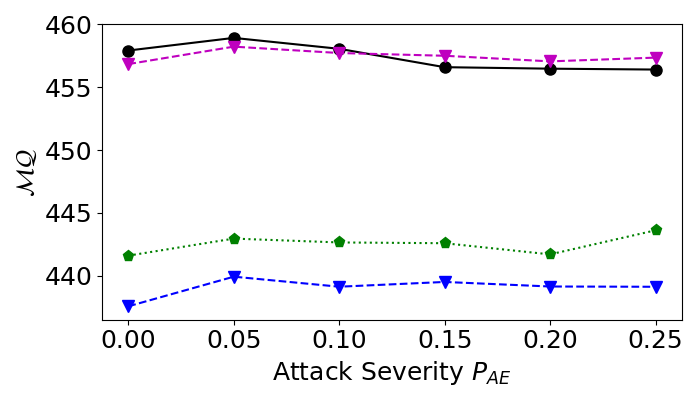}}
  \subfigure[\footnotesize{Remaining Energy ($\mathcal{RE}$)}]{
    \includegraphics[width=0.43\textwidth, height=0.3\textwidth]{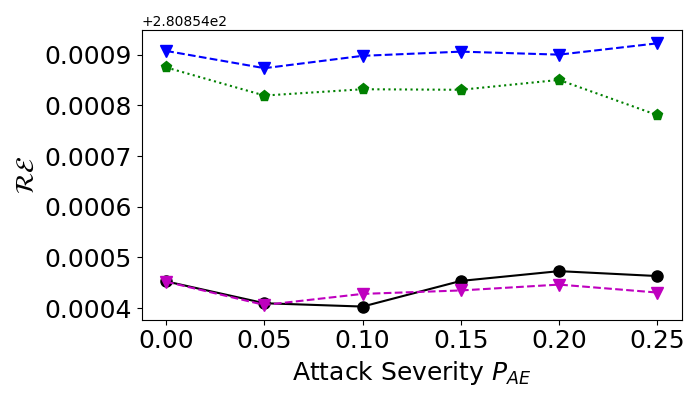}}
    \subfigure[\footnotesize{Run Time ($R_T$)}]{
    \includegraphics[width=0.43\textwidth, height=0.3\textwidth]{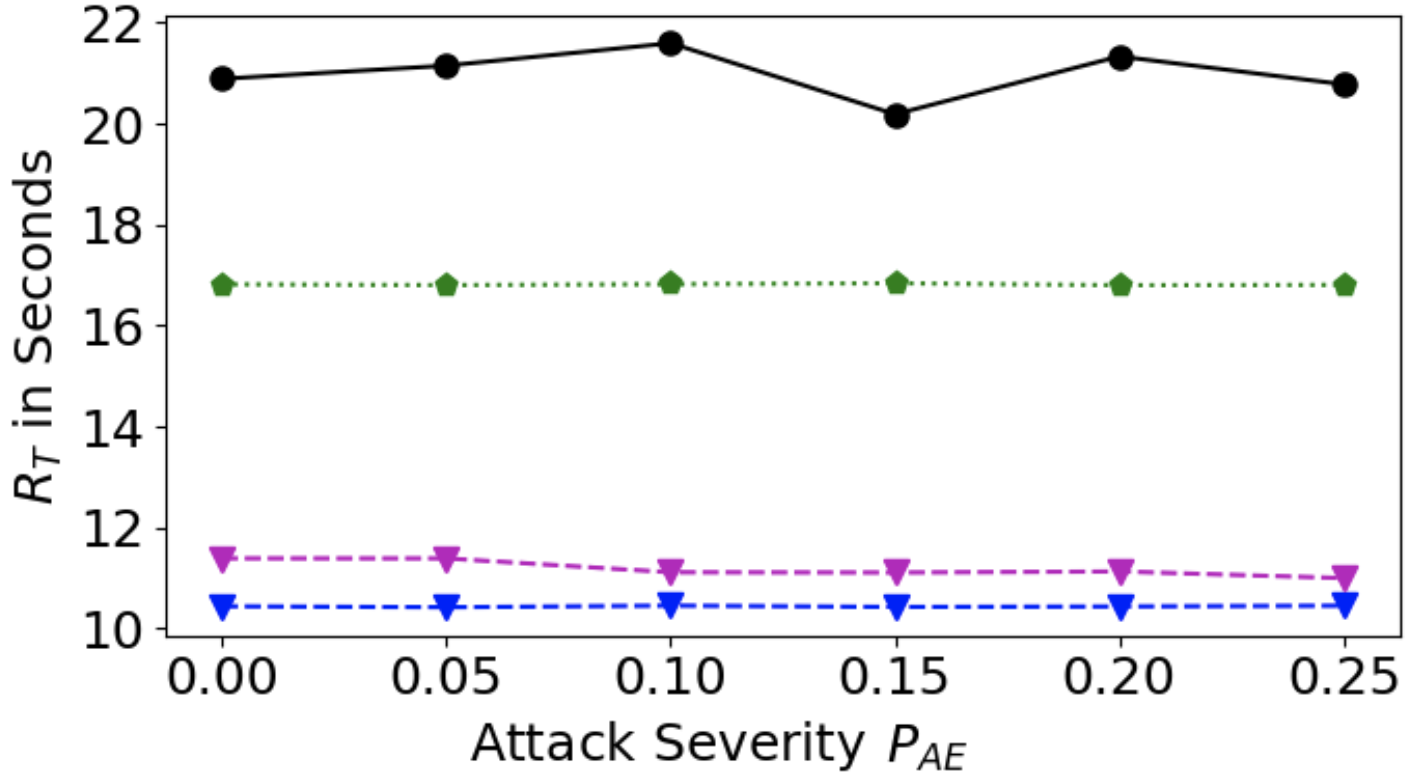}}
    
    \caption{Effect of Varying Adversarial Example Severity ($P_{AE}$) on Gateways.}
\label{fig:nn}
\end{figure*}

\begin{figure*}[!ht]
  \centering
  \subfigure{
    \includegraphics[width=0.7\textwidth, height=0.035\textwidth]{figs/sen_a/legend_both.png}}
    \setcounter{subfigure}{0}
    \vspace{-2mm}
    
  \subfigure[\footnotesize{Accumulated Reward ($\mathcal{R}$)}]{
    \includegraphics[width=0.43\textwidth, height=0.3\textwidth]{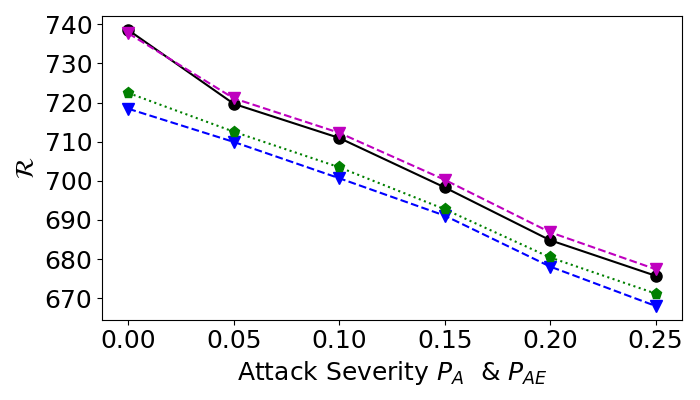}}
  \subfigure[\footnotesize{Monitoring Quality ($\mathcal{MQ}$)}]{
    \includegraphics[width=0.43\textwidth, height=0.3\textwidth]{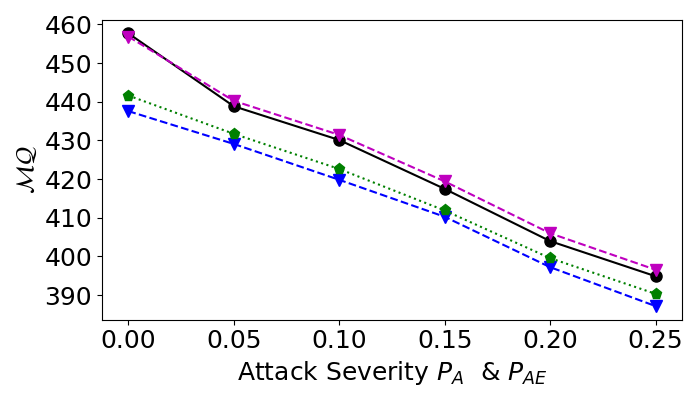}}
  \subfigure[\footnotesize{Remaining Energy ($\mathcal{RE}$)}]{
    \includegraphics[width=0.43\textwidth, height=0.3\textwidth]{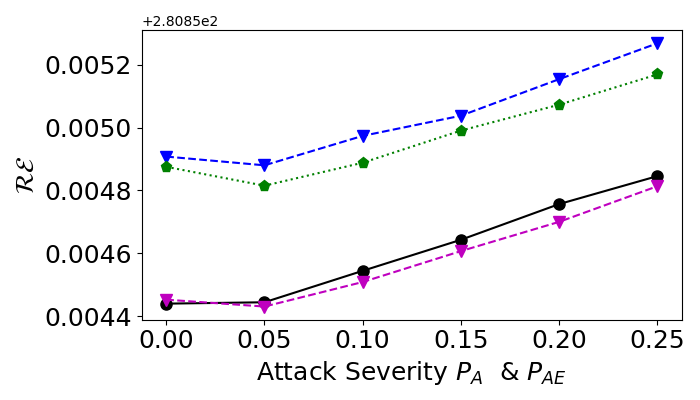}}
    \subfigure[\footnotesize{Run Time ($R_T$)}]{
    \includegraphics[width=0.43\textwidth, height=0.3\textwidth]{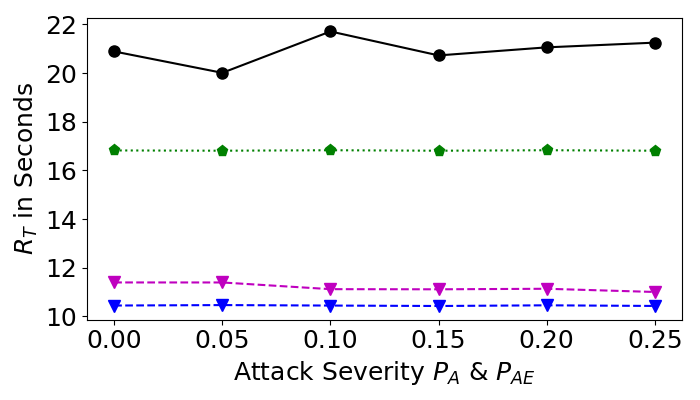}}
    
    \caption{Effect of Varying Cyber Attack ($P_A$) and Adversarial Example ($P_{AE}$) Severity.}
\label{fig:both}
\vspace{-4mm}
\end{figure*}

In our analysis, we derive the data points for comparison from the average values obtained across the initial 50 episodes of the training curves. This methodology ensures a meaningful and standardized comparison framework, which is especially relevant since all PPO-based strategies eventually converge to a comparable performance level. This approach allows us to highlight the efficiency and effectiveness of various methods during the critical early phase of training, providing insights into their initial adaptation and learning speeds.

\subsubsection{\bf Effect of Varying Cyber Attack Severity ($P_A$) on Sensors} Fig.~\ref{fig:da} shows how varying degrees of attack severity ($P_A$) on sensor nodes influence key performance indicators. As $P_A$ increases, the monitoring quality ($\mathcal{MQ}$) diminishes, primarily due to the augmented influx of compromised data, undermining the integrity of the monitoring process. Simultaneously, a higher $P_A$ leads to an increase in the remaining energy ($\mathcal{RE}$) among all schemes. This is attributed to the reduced incentive for data transmission amidst a higher prevalence of compromised data, thus conserving energy that would otherwise be expended in sending data. 

In our study, we observe that the performance of our proposed DT-PPO scheme aligns closely with that of the TL-PPO-FT. This finding supports our comparative analysis, demonstrating that DT-PPO can achieve equivalent performance levels during the initial stages of the simulation, comparable to a PPO model that has undergone a complete transfer process. This equivalence in early performance underscores the efficiency of the DT-PPO approach in quickly adapting to and addressing the simulation's requirements, highlighting its potential as a robust alternative to fully transferred models in early training phases.  The hierarchy of performance across different $P_A$ levels aligns as follows: TL-PPO-FT $\approx$ DT-PPO $\geq$ TL-PPO-PT $\approx$ PPO $\geq$ AED $\geq$ Random $\geq$ FE.  This ordering reflects the comparative advantage of PPO-enhanced schemes in maintaining higher monitoring quality and energy efficiency under adversarial conditions. The adaptability and learning efficiency of DT-PPO and TL-PPO-FT, in particular, stand out, indicating their resilience in facing the dual challenge of sustaining system performance while managing energy resources effectively in the face of cybersecurity threats.


\subsubsection{\bf Effect of Varying Adversarial Example Severity ($P_{AE}$) on Gateways}Fig.~\ref{fig:nn} showcases how the severity of adversarial examples ($P_{AE}$) impacts the performance metrics of gateways. The findings reveal that the DT-PPO scheme exhibits greater resilience to adversarial attacks than other PPO strategies.  The enhanced resilience of the DT-PPO approach can be attributed to its integration of DT strategies, which rely heavily on utility values for selecting the optimal actions at the onset of the simulation. This foundational reliance on utility values rather than direct environmental feedback makes the DT-PPO scheme inherently more robust to adversarial attacks aimed at exploiting vulnerabilities in the PPO algorithm. As the severity of these adversarial attacks, $P_{AE}$, increases, the DT-PPO's unique mechanism of action selection based on pre-determined utility values rather than learned policies, enables it to maintain performance where traditional PPO-based methods, including TL-PPO-FT, might falter.  Consequently, as adversarial conditions intensify, DT-PPO's strategic advantage becomes more pronounced, allowing it to outperform TL-PPO-FT eventually. This demonstrates the DT-PPO scheme's superior durability and adaptability in the face of escalating adversarial threats.  Furthermore, the analysis indicates that the overall system's vulnerability to adversarial attacks does not markedly increase with the level of attack severity. This resilience highlights the effectiveness of incorporating the proposed DT-guided techniques in PPO frameworks to enhance their robustness and adaptability in adversarial environments.


\subsubsection{\bf Effect of Varying Cyber Attack ($P_A$) and Adversarial Example ($P_{AE}$) Severity}  Fig.~\ref{fig:both} elucidates the dual impact of varying both $P_A$ on sensor nodes and $P_{AE}$ on gateways, on the performance metrics of DRL-based monitoring schemes. Our results show that the DT-PPO scheme marginally surpasses TL-PPO-FT in accumulated rewards as the severity of attacks escalates. This underscores the efficacy of the DT-PPO model in adjusting to adversarial conditions. Given their foundation in the PPO-based DRL framework, both strategies can formulate similar policies. However, the incorporation of DT-guided methods within DT-PPO plays a crucial role in bolstering its resilience to adversarial attacks, underscoring the significant impact of decision theory techniques on enhancing adaptability and robustness against such challenges. 

When considering the combined effects of $P_A$ and $P_{AE}$, the performance hierarchy of the schemes is established as DT-PPO $\approx$ TL-PPO-FT $\geq$ TL-PPO-PT $\geq$ PPO. This ranking underscores the superior adaptability and resilience of fully transferred PPO and DT-guided PPO (TL-PPO-FT and DT-PPO) over their counterparts, including standalone PPO and partially transferred PPO variations. The comparative resilience of these schemes to adversarial pressures, both from compromised sensor data and gateway-level neural network manipulations, showcases the critical role of advanced DT-guided strategies in fortifying DRL-based systems against a spectrum of cyber threats.  The algorithmic runtime remains unaffected by variations in attack severity for both $P_{A}$ and $P_{AE}$.  This lack of sensitivity arises because the operations associated with these attacks do not influence the computational duration of the algorithm.

\section{Conclusions \& Future Work} 
\label{sec:conclusions-future-work}

This work proposed a novel attack-resilient and energy-adaptive monitoring system for smart farms, integrating multi-agent deep reinforcement learning (DRL) with decision theory (DT) to optimize monitoring quality while ensuring energy efficiency. By leveraging DT-guided strategies, we mitigated DRL's cold start problem and enhanced adaptability in adversarial environments. Our study provides new insights into balancing energy constraints and resilience against cyber threats in IoT-driven smart farming applications.

\subsection{Key Findings}
\begin{itemize}
    \item The DT-guided DRL approach effectively enhances monitoring quality while maintaining energy efficiency, demonstrating a significant improvement in learning speed and initial performance compared to conventional DRL.
    \item The DT-PPO (Decision Theory - Proximal Policy Optimization) scheme performs better in optimizing monitoring quality and sensor energy conservation, outperforming transfer learning-based methods when pre-trained models are unavailable.
    \item Transfer learning (TL) within DRL reduces training duration when an adequately pre-trained model is available, accelerating agent learning in well-studied environments.
    \item Sensitivity analyses revealed that DT-PPO demonstrates robustness to cyber-attacks and adversarial manipulations, maintaining stable performance under increasing attack severity.
    \item The proposed uncertainty-aware aggregation method enhances system resilience by filtering deceptive data, improving decision-making in adversarial settings.
\end{itemize}

\subsection{Future Work}
Several promising directions for future research remain:

\begin{itemize}
    \item \textit{Scaling to large-scale deployments:} We will explore integrating additional DRL agents and expanding the system to manage larger sensor networks efficiently. This includes optimizing multi-agent coordination for scalability and robustness in extensive farming environments.
    \item \textit{Extending DT-guided DRL to complex tasks:} While this study focused on constrained environments such as smart farms, we plan to extend the DT-guided DRL framework to more complex and dynamic real-world applications, such as autonomous agricultural machinery, livestock tracking across larger geographic areas, and multi-robot coordination.
    \item \textit{Real-world implementation and validation:} The proposed framework will be deployed and evaluated in real-world smart farm testbeds to assess its practical applicability under real environmental conditions, network constraints, and unpredictable adversarial attacks.
    \item \textit{Enhancing adaptive security mechanisms:} Future work will explore integrating adversarial defense mechanisms, such as adversarial training and anomaly detection, to further improve robustness against cyber threats targeting DRL-based control systems.
    \item \textit{Adaptive and context-aware decision-making:} We will consider incorporating adaptive decision models that adapt to specific farm conditions, livestock behaviors, and dynamic energy harvesting capabilities, enabling a more tailored and efficient monitoring framework.
\end{itemize}

By addressing these challenges, our work aims to pave the way for the next generation of secure, energy-efficient, and intelligent smart farming systems, enhancing both resilience and sustainability in agricultural IoT applications.

\bibliographystyle{ACM-Reference-Format}
\bibliography{ref}


\begin{thebibliography}{40}


\ifx \showCODEN    \undefined \def \showCODEN     #1{\unskip}     \fi
\ifx \showDOI      \undefined \def \showDOI       #1{#1}\fi
\ifx \showISBNx    \undefined \def \showISBNx     #1{\unskip}     \fi
\ifx \showISBNxiii \undefined \def \showISBNxiii  #1{\unskip}     \fi
\ifx \showISSN     \undefined \def \showISSN      #1{\unskip}     \fi
\ifx \showLCCN     \undefined \def \showLCCN      #1{\unskip}     \fi
\ifx \shownote     \undefined \def \shownote      #1{#1}          \fi
\ifx \showarticletitle \undefined \def \showarticletitle #1{#1}   \fi
\ifx \showURL      \undefined \def \showURL       {\relax}        \fi
\providecommand\bibfield[2]{#2}
\providecommand\bibinfo[2]{#2}
\providecommand\natexlab[1]{#1}
\providecommand\showeprint[2][]{arXiv:#2}

\bibitem[Aerts et~al\mbox{.}(2023)]%
        {aerts2023bayesian}
\bibfield{author}{\bibinfo{person}{Frederik Aerts}, \bibinfo{person}{Luca
  Lanzilao}, {and} \bibinfo{person}{Johan Meyers}.}
  \bibinfo{year}{2023}\natexlab{}.
\newblock \showarticletitle{Bayesian uncertainty quantification framework for
  wake model calibration and validation with historical wind farm power data}.
\newblock \bibinfo{journal}{\emph{Wind Energy}} \bibinfo{volume}{26},
  \bibinfo{number}{8} (\bibinfo{year}{2023}), \bibinfo{pages}{786--802}.
\newblock


\bibitem[Alemayehu and Kim(2017)]%
        {alemayehu17}
\bibfield{author}{\bibinfo{person}{Temesgen~Seyoum Alemayehu} {and}
  \bibinfo{person}{Jai-Hoon Kim}.} \bibinfo{year}{2017}\natexlab{}.
\newblock \showarticletitle{Efficient nearest neighbor heuristic {TSP}
  algorithms for reducing data acquisition latency of {UAV} relay {WSN}}.
\newblock \bibinfo{journal}{\emph{Wireless Personal Communications}}
  \bibinfo{volume}{95} (\bibinfo{year}{2017}), \bibinfo{pages}{3271--3285}.
\newblock


\bibitem[Aliyu and Liu(2023)]%
        {aliyu2023blockchain}
\bibfield{author}{\bibinfo{person}{Ahmed~Abubakar Aliyu} {and}
  \bibinfo{person}{Jinshuo Liu}.} \bibinfo{year}{2023}\natexlab{}.
\newblock \showarticletitle{Blockchain-Based Smart Farm Security Framework for
  the Internet of Things}.
\newblock \bibinfo{journal}{\emph{Sensors}} \bibinfo{volume}{23},
  \bibinfo{number}{18} (\bibinfo{year}{2023}), \bibinfo{pages}{7992}.
\newblock


\bibitem[Arulkumaran et~al\mbox{.}(2017)]%
        {kai2017}
\bibfield{author}{\bibinfo{person}{Kai Arulkumaran},
  \bibinfo{person}{Marc~Peter Deisenroth}, \bibinfo{person}{Miles Brundage},
  {and} \bibinfo{person}{Anil~Anthony Bharath}.}
  \bibinfo{year}{2017}\natexlab{}.
\newblock \showarticletitle{Deep Reinforcement Learning: A Brief Survey}.
\newblock \bibinfo{journal}{\emph{IEEE Signal Processing Magazine}}
  \bibinfo{volume}{34}, \bibinfo{number}{6} (\bibinfo{year}{2017}),
  \bibinfo{pages}{26--38}.
\newblock
\urldef\tempurl%
\url{https://doi.org/10.1109/MSP.2017.2743240}
\showDOI{\tempurl}


\bibitem[Ayas et~al\mbox{.}(2022)]%
        {ayas22}
\bibfield{author}{\bibinfo{person}{Mustafa~Sinasi Ayas}, \bibinfo{person}{Selen
  Ayas}, {and} \bibinfo{person}{Seddik~M. Djouadi}.}
  \bibinfo{year}{2022}\natexlab{}.
\newblock \showarticletitle{Projected Gradient Descent Adversarial Attack and
  Its Defense on a Fault Diagnosis System}. In \bibinfo{booktitle}{\emph{2022
  45th International Conference on Telecommunications and Signal Processing
  (TSP)}}. \bibinfo{pages}{36--39}.
\newblock
\urldef\tempurl%
\url{https://doi.org/10.1109/TSP55681.2022.9851334}
\showDOI{\tempurl}


\bibitem[Bogue(2012)]%
        {bogue2012solar}
\bibfield{author}{\bibinfo{person}{Robert Bogue}.}
  \bibinfo{year}{2012}\natexlab{}.
\newblock \showarticletitle{Solar-powered sensors: a review of products and
  applications}.
\newblock \bibinfo{journal}{\emph{Sensor Review}} \bibinfo{volume}{32},
  \bibinfo{number}{2} (\bibinfo{year}{2012}), \bibinfo{pages}{95--100}.
\newblock


\bibitem[Chae and Cho(2018)]%
        {chae2018enhanced}
\bibfield{author}{\bibinfo{person}{Cheol-Joo Chae} {and}
  \bibinfo{person}{Han-Jin Cho}.} \bibinfo{year}{2018}\natexlab{}.
\newblock \showarticletitle{Enhanced secure device authentication algorithm in
  P2P-based smart farm system}.
\newblock \bibinfo{journal}{\emph{Peer-to-peer networking and applications}}
  \bibinfo{volume}{11} (\bibinfo{year}{2018}), \bibinfo{pages}{1230--1239}.
\newblock


\bibitem[Cutler and How(2015)]%
        {cutler2015efficient}
\bibfield{author}{\bibinfo{person}{Mark Cutler} {and}
  \bibinfo{person}{Jonathan~P How}.} \bibinfo{year}{2015}\natexlab{}.
\newblock \showarticletitle{Efficient reinforcement learning for robots using
  informative simulated priors}. In \bibinfo{booktitle}{\emph{2015 IEEE
  international conference on robotics and automation (ICRA)}}. IEEE,
  \bibinfo{pages}{2605--2612}.
\newblock


\bibitem[dos Reis et~al\mbox{.}(2021)]%
        {dos2021}
\bibfield{author}{\bibinfo{person}{B~R dos Reis}, \bibinfo{person}{Z Easton},
  \bibinfo{person}{R~R White}, {and} \bibinfo{person}{D Fuka}.}
  \bibinfo{year}{2021}\natexlab{}.
\newblock \showarticletitle{A {LoRa} sensor network for monitoring pastured
  livestock location and activity}.
\newblock \bibinfo{journal}{\emph{Translational Animal Science}}
  \bibinfo{volume}{5}, \bibinfo{number}{2} (\bibinfo{date}{Jan.}
  \bibinfo{year}{2021}), \bibinfo{pages}{1--9}.
\newblock


\bibitem[Eichner et~al\mbox{.}(2023)]%
        {eichner2023optimal}
\bibfield{author}{\bibinfo{person}{Lukas Eichner}, \bibinfo{person}{Ronald
  Schneider}, {and} \bibinfo{person}{Matthias Bae{\ss}ler}.}
  \bibinfo{year}{2023}\natexlab{}.
\newblock \showarticletitle{Optimal vibration sensor placement for jacket
  support structures of offshore wind turbines based on value of information
  analysis}.
\newblock \bibinfo{journal}{\emph{Ocean Engineering}}  \bibinfo{volume}{288}
  (\bibinfo{year}{2023}), \bibinfo{pages}{115407}.
\newblock


\bibitem[Giordano et~al\mbox{.}(2023)]%
        {giordano2023value}
\bibfield{author}{\bibinfo{person}{Pier~Francesco Giordano},
  \bibinfo{person}{Leandro Iannacone}, {and} \bibinfo{person}{Maria~Pina
  Limongelli}.} \bibinfo{year}{2023}\natexlab{}.
\newblock \showarticletitle{Value of Seismic Structural Health Monitoring
  Information for Management of Civil Structures Under Different Prior
  Knowledge Scenarios}. In \bibinfo{booktitle}{\emph{International Conference
  on Experimental Vibration Analysis for Civil Engineering Structures}}.
  Springer, \bibinfo{pages}{11--20}.
\newblock


\bibitem[Goodfellow et~al\mbox{.}(2015)]%
        {fel14}
\bibfield{author}{\bibinfo{person}{Ian Goodfellow}, \bibinfo{person}{Jonathon
  Shlens}, {and} \bibinfo{person}{Christian Szegedy}.}
  \bibinfo{year}{2015}\natexlab{}.
\newblock \showarticletitle{Explaining and Harnessing Adversarial Examples}. In
  \bibinfo{booktitle}{\emph{International Conference on Learning
  Representations (ICLR)}}.
\newblock


\bibitem[Gupta et~al\mbox{.}(2020)]%
        {gupta20}
\bibfield{author}{\bibinfo{person}{Maanak Gupta}, \bibinfo{person}{Mahmoud
  Abdelsalam}, \bibinfo{person}{Sajad Khorsandroo}, {and}
  \bibinfo{person}{Sudip Mittal}.} \bibinfo{year}{2020}\natexlab{}.
\newblock \showarticletitle{Security and privacy in smart farming: Challenges
  and opportunities}.
\newblock \bibinfo{journal}{\emph{IEEE Access}}  \bibinfo{volume}{8}
  (\bibinfo{year}{2020}), \bibinfo{pages}{34564--34584}.
\newblock


\bibitem[J\o{}sang(2016)]%
        {sj16}
\bibfield{author}{\bibinfo{person}{Audun J\o{}sang}.}
  \bibinfo{year}{2016}\natexlab{}.
\newblock \bibinfo{booktitle}{\emph{Subjective Logic: A Formalism for Reasoning
  Under Uncertainty} (\bibinfo{edition}{1st} ed.)}.
\newblock \bibinfo{publisher}{Springer Publishing Company, Incorporated}.
\newblock
\showISBNx{3319423355}


\bibitem[Kang et~al\mbox{.}(2019)]%
        {kang2019generalization}
\bibfield{author}{\bibinfo{person}{Katie Kang}, \bibinfo{person}{Suneel
  Belkhale}, \bibinfo{person}{Gregory Kahn}, \bibinfo{person}{Pieter Abbeel},
  {and} \bibinfo{person}{Sergey Levine}.} \bibinfo{year}{2019}\natexlab{}.
\newblock \showarticletitle{Generalization through simulation: Integrating
  simulated and real data into deep reinforcement learning for vision-based
  autonomous flight}. In \bibinfo{booktitle}{\emph{2019 international
  conference on robotics and automation (ICRA)}}. IEEE,
  \bibinfo{pages}{6008--6014}.
\newblock


\bibitem[Kim et~al\mbox{.}(2014)]%
        {kim2014ambient}
\bibfield{author}{\bibinfo{person}{Sangkil Kim}, \bibinfo{person}{Rushi Vyas},
  \bibinfo{person}{Jo Bito}, \bibinfo{person}{Kyriaki Niotaki},
  \bibinfo{person}{Ana Collado}, \bibinfo{person}{Apostolos Georgiadis}, {and}
  \bibinfo{person}{Manos~M Tentzeris}.} \bibinfo{year}{2014}\natexlab{}.
\newblock \showarticletitle{Ambient {RF} energy-harvesting technologies for
  self-sustainable standalone wireless sensor platforms}.
\newblock \bibinfo{journal}{\emph{Proc. IEEE}} \bibinfo{volume}{102},
  \bibinfo{number}{11} (\bibinfo{year}{2014}), \bibinfo{pages}{1649--1666}.
\newblock


\bibitem[Kumar et~al\mbox{.}(2019)]%
        {kumar19}
\bibfield{author}{\bibinfo{person}{Siddhant Kumar}, \bibinfo{person}{Gourav
  Chowdhary}, \bibinfo{person}{Venkanna Udutalapally},
  \bibinfo{person}{Debanjan Das}, {and} \bibinfo{person}{Saraju~P. Mohanty}.}
  \bibinfo{year}{2019}\natexlab{}.
\newblock \showarticletitle{{gCrop}: {Internet-of-Leaf-Things (IoLT)} for
  Monitoring of the Growth of Crops in Smart Agriculture}. In
  \bibinfo{booktitle}{\emph{2019 IEEE International Symposium on Smart
  Electronic Systems (iSES) (Formerly iNiS)}}. \bibinfo{pages}{53--56}.
\newblock
\urldef\tempurl%
\url{https://doi.org/10.1109/iSES47678.2019.00024}
\showDOI{\tempurl}


\bibitem[Kuntke et~al\mbox{.}(2022)]%
        {kuntke2022lorawan}
\bibfield{author}{\bibinfo{person}{Franz Kuntke}, \bibinfo{person}{Vladimir
  Romanenko}, \bibinfo{person}{Sebastian Linsner}, \bibinfo{person}{Enno
  Steinbrink}, {and} \bibinfo{person}{Christian Reuter}.}
  \bibinfo{year}{2022}\natexlab{}.
\newblock \showarticletitle{{LoRaWAN} security issues and mitigation options by
  the example of agricultural {IoT} scenarios}.
\newblock \bibinfo{journal}{\emph{Transactions on Emerging Telecommunications
  Technologies}} \bibinfo{volume}{33}, \bibinfo{number}{5}
  (\bibinfo{year}{2022}), \bibinfo{pages}{e4452}.
\newblock


\bibitem[Liu et~al\mbox{.}(2012)]%
        {liu12}
\bibfield{author}{\bibinfo{person}{Min Liu}, \bibinfo{person}{Shijun Xu}, {and}
  \bibinfo{person}{Siyi Sun}.} \bibinfo{year}{2012}\natexlab{}.
\newblock \showarticletitle{An agent-assisted {QoS}-based routing algorithm for
  wireless sensor networks}.
\newblock \bibinfo{journal}{\emph{Journal of Network and Computer
  Applications}} \bibinfo{volume}{35}, \bibinfo{number}{1}
  (\bibinfo{year}{2012}), \bibinfo{pages}{29--36}.
\newblock


\bibitem[Liu et~al\mbox{.}(2018)]%
        {liu18-Purdue}
\bibfield{author}{\bibinfo{person}{Yingqi Liu}, \bibinfo{person}{Shiqing Ma},
  \bibinfo{person}{Yousra Aafer}, \bibinfo{person}{Wen-Chuan Lee},
  \bibinfo{person}{Juan Zhai}, \bibinfo{person}{Weihang Wang}, {and}
  \bibinfo{person}{Xiangyu Zhang}.} \bibinfo{year}{2018}\natexlab{}.
\newblock \showarticletitle{Trojaning Attack on Neural Networks}. In
  \bibinfo{booktitle}{\emph{Network and Distributed System Security Symposium
  (NDSS)}}. \bibinfo{publisher}{Internet Society}.
\newblock


\bibitem[Mathuna et~al\mbox{.}(2008)]%
        {mathuna2008energy}
\bibfield{author}{\bibinfo{person}{Cian~O Mathuna}, \bibinfo{person}{Terence
  O’Donnell}, \bibinfo{person}{Rafael~V Martinez-Catala},
  \bibinfo{person}{James Rohan}, {and} \bibinfo{person}{Brendan O’Flynn}.}
  \bibinfo{year}{2008}\natexlab{}.
\newblock \showarticletitle{Energy scavenging for long-term deployable wireless
  sensor networks}.
\newblock \bibinfo{journal}{\emph{Talanta}} \bibinfo{volume}{75},
  \bibinfo{number}{3} (\bibinfo{year}{2008}), \bibinfo{pages}{613--623}.
\newblock


\bibitem[Metallidou et~al\mbox{.}(2020)]%
        {metallidou2020energy}
\bibfield{author}{\bibinfo{person}{Chrysi~K Metallidou},
  \bibinfo{person}{Kostas~E Psannis}, {and}
  \bibinfo{person}{Eugenia~Alexandropoulou Egyptiadou}.}
  \bibinfo{year}{2020}\natexlab{}.
\newblock \showarticletitle{Energy efficiency in smart buildings: {IoT}
  approaches}.
\newblock \bibinfo{journal}{\emph{IEEE Access}}  \bibinfo{volume}{8}
  (\bibinfo{year}{2020}), \bibinfo{pages}{63679--63699}.
\newblock


\bibitem[Microchip(2018)]%
        {manualR2}
Microchip \bibinfo{year}{2018}\natexlab{}.
\newblock \bibinfo{booktitle}{\emph{\uppercase{SAM} R34$/$R35 Low Power
  Lo\uppercase{R}a\textregistered{} Sub-\uppercase{GH}z Si\uppercase{P}
  Datasheet}}.
\newblock Microchip.
\newblock
\urldef\tempurl%
\url{http://ww1.microchip.com/downloads/en/DeviceDoc/SAMR34-R35-Low-Power-LoRa-Sub-GHz-SiP-Data-Sheet-DS70005356B.pdf}
\showURL{%
\tempurl}


\bibitem[Nefedov and Fil(2023)]%
        {nefedov2023model}
\bibfield{author}{\bibinfo{person}{Leonid Nefedov} {and}
  \bibinfo{person}{Nataliia Fil}.} \bibinfo{year}{2023}\natexlab{}.
\newblock \showarticletitle{The model of the regional environmental monitoring
  system organization}. In \bibinfo{booktitle}{\emph{The IEEE 13th
  International Conference on Dependable Systems, Services and Technologies
  (DESSERT)}}. \bibinfo{pages}{1--6}.
\newblock


\bibitem[Nguyen et~al\mbox{.}(2021)]%
        {nguyen2021federated}
\bibfield{author}{\bibinfo{person}{Tri~Gia Nguyen}, \bibinfo{person}{Trung~V
  Phan}, \bibinfo{person}{Dinh~Thai Hoang}, \bibinfo{person}{Tu~N Nguyen},
  {and} \bibinfo{person}{Chakchai So-In}.} \bibinfo{year}{2021}\natexlab{}.
\newblock \showarticletitle{Federated deep reinforcement learning for traffic
  monitoring in {SDN}-based {IoT} networks}.
\newblock \bibinfo{journal}{\emph{IEEE Trans. Cognitive Communications and
  Networking}} \bibinfo{volume}{7}, \bibinfo{number}{4} (\bibinfo{year}{2021}),
  \bibinfo{pages}{1048--1065}.
\newblock


\bibitem[Nguyen et~al\mbox{.}(2023)]%
        {nguyen2023utilizing}
\bibfield{author}{\bibinfo{person}{Thu~Nga Nguyen}, \bibinfo{person}{Trong~Binh
  Nguyen}, \bibinfo{person}{Trinh Van~Chien}, {and} \bibinfo{person}{Tien~Hoa
  Nguyen}.} \bibinfo{year}{2023}\natexlab{}.
\newblock \showarticletitle{Utilizing Deep Reinforcement Learning to Control
  {UAV} Movement for Environmental Monitoring}.
\newblock \bibinfo{journal}{\emph{International Journal of Electrical and
  Electronic Engineering \& Telecommunications}} \bibinfo{volume}{12},
  \bibinfo{number}{5} (\bibinfo{year}{2023}), \bibinfo{pages}{317--325}.
\newblock


\bibitem[North(1968)]%
        {north1968tutorial}
\bibfield{author}{\bibinfo{person}{D~Warner North}.}
  \bibinfo{year}{1968}\natexlab{}.
\newblock \showarticletitle{A Tutorial Introduction to Decision Theory}.
\newblock \bibinfo{journal}{\emph{IEEE Trans. Systems Science and Cybernetics}}
  \bibinfo{volume}{4}, \bibinfo{number}{3} (\bibinfo{year}{1968}),
  \bibinfo{pages}{200--210}.
\newblock


\bibitem[Osi{\'n}ski et~al\mbox{.}(2020)]%
        {osinski2020simulation}
\bibfield{author}{\bibinfo{person}{B{\l}a{\.z}ej Osi{\'n}ski},
  \bibinfo{person}{Adam Jakubowski}, \bibinfo{person}{Pawe{\l} Zi{\k{e}}cina},
  \bibinfo{person}{Piotr Mi{\l}o{\'s}}, \bibinfo{person}{Christopher Galias},
  \bibinfo{person}{Silviu Homoceanu}, {and} \bibinfo{person}{Henryk
  Michalewski}.} \bibinfo{year}{2020}\natexlab{}.
\newblock \showarticletitle{Simulation-based reinforcement learning for
  real-world autonomous driving}. In \bibinfo{booktitle}{\emph{2020 IEEE
  international conference on robotics and automation (ICRA)}}. IEEE,
  \bibinfo{pages}{6411--6418}.
\newblock


\bibitem[Parmigiani and Inoue(2009)]%
        {parmigiani2009decision}
\bibfield{author}{\bibinfo{person}{Giovanni Parmigiani} {and}
  \bibinfo{person}{Lurdes Inoue}.} \bibinfo{year}{2009}\natexlab{}.
\newblock \bibinfo{booktitle}{\emph{Decision theory: Principles and
  approaches}}.
\newblock \bibinfo{publisher}{John Wiley \& Sons}.
\newblock


\bibitem[Peterson(2017)]%
        {peterson2017introduction}
\bibfield{author}{\bibinfo{person}{Martin Peterson}.}
  \bibinfo{year}{2017}\natexlab{}.
\newblock \bibinfo{booktitle}{\emph{An Introduction to Decision Theory}}.
\newblock \bibinfo{publisher}{Cambridge University Press}.
\newblock


\bibitem[Saheed and Arowolo(2021)]%
        {saheed21}
\bibfield{author}{\bibinfo{person}{Yakub~Kayode Saheed} {and}
  \bibinfo{person}{Micheal~Olaolu Arowolo}.} \bibinfo{year}{2021}\natexlab{}.
\newblock \showarticletitle{Efficient Cyber Attack Detection on the Internet of
  Medical Things-Smart Environment Based on Deep Recurrent Neural Network and
  Machine Learning Algorithms}.
\newblock \bibinfo{journal}{\emph{IEEE Access}}  \bibinfo{volume}{9}
  (\bibinfo{year}{2021}), \bibinfo{pages}{161546--161554}.
\newblock
\urldef\tempurl%
\url{https://doi.org/10.1109/ACCESS.2021.3128837}
\showDOI{\tempurl}


\bibitem[{Schulman, et al.}(2017)]%
        {schu17}
\bibfield{author}{\bibinfo{person}{John {Schulman, et al.}}}
  \bibinfo{year}{2017}\natexlab{}.
\newblock \showarticletitle{Proximal policy optimization algorithms}.
\newblock \bibinfo{journal}{\emph{arXiv preprint arXiv:1707.06347}}
  (\bibinfo{year}{2017}).
\newblock


\bibitem[Sontowski et~al\mbox{.}(2020)]%
        {sontowski2020}
\bibfield{author}{\bibinfo{person}{Sina Sontowski}, \bibinfo{person}{Maanak
  Gupta}, \bibinfo{person}{Sai~Sree Laya~Chukkapalli}, \bibinfo{person}{Mahmoud
  Abdelsalam}, \bibinfo{person}{Sudip Mittal}, \bibinfo{person}{Anupam Joshi},
  {and} \bibinfo{person}{Ravi Sandhu}.} \bibinfo{year}{2020}\natexlab{}.
\newblock \showarticletitle{Cyber Attacks on Smart Farming Infrastructure}. In
  \bibinfo{booktitle}{\emph{2020 IEEE 6th International Conference on
  Collaboration and Internet Computing (CIC)}}. \bibinfo{pages}{135--143}.
\newblock
\urldef\tempurl%
\url{https://doi.org/10.1109/CIC50333.2020.00025}
\showDOI{\tempurl}


\bibitem[Sultan et~al\mbox{.}(2021)]%
        {sultan2021energy}
\bibfield{author}{\bibinfo{person}{Salman~Md Sultan}, \bibinfo{person}{Muhammad
  Waleed}, \bibinfo{person}{Jae-Young Pyun}, {and} \bibinfo{person}{Tai-Won
  Um}.} \bibinfo{year}{2021}\natexlab{}.
\newblock \showarticletitle{Energy conservation for Internet of Things tracking
  applications using deep reinforcement learning}.
\newblock \bibinfo{journal}{\emph{Sensors}} \bibinfo{volume}{21},
  \bibinfo{number}{9} (\bibinfo{year}{2021}), \bibinfo{pages}{3261}.
\newblock


\bibitem[Texas Instruments(2016)]%
        {manualR3}
Texas Instruments \bibinfo{year}{2016}\natexlab{}.
\newblock \bibinfo{booktitle}{\emph{CC2640R2F SimpleLink\texttrademark{}
  Bluetooth\textregistered{} 5.1 Low Energy Wireless \uppercase{MCU}}}.
\newblock Texas Instruments.
\newblock
\urldef\tempurl%
\url{https://www.ti.com/product/CC2640R2F}
\showURL{%
\tempurl}
\newblock
\shownote{Rev. C}.


\bibitem[Vangala et~al\mbox{.}(2021)]%
        {vangala2021smart}
\bibfield{author}{\bibinfo{person}{Anusha Vangala}, \bibinfo{person}{Anil~Kumar
  Sutrala}, \bibinfo{person}{Ashok~Kumar Das}, {and} \bibinfo{person}{Minho
  Jo}.} \bibinfo{year}{2021}\natexlab{}.
\newblock \showarticletitle{Smart contract-based blockchain-envisioned
  authentication scheme for smart farming}.
\newblock \bibinfo{journal}{\emph{IEEE Internet of Things Journal}}
  \bibinfo{volume}{8}, \bibinfo{number}{13} (\bibinfo{year}{2021}),
  \bibinfo{pages}{10792--10806}.
\newblock


\bibitem[Wang et~al\mbox{.}(2020)]%
        {wang2020slora}
\bibfield{author}{\bibinfo{person}{Xiong Wang}, \bibinfo{person}{Linghe Kong},
  \bibinfo{person}{Zucheng Wu}, \bibinfo{person}{Long Cheng},
  \bibinfo{person}{Chenren Xu}, {and} \bibinfo{person}{Guihai Chen}.}
  \bibinfo{year}{2020}\natexlab{}.
\newblock \showarticletitle{{SLoRa}: Towards secure {LoRa} communications with
  fine-grained physical layer features}. In
  \bibinfo{booktitle}{\emph{Proceedings of the 18th Conference on Embedded
  Networked Sensor Systems}}. \bibinfo{pages}{258--270}.
\newblock


\bibitem[Yazdinejad et~al\mbox{.}(2021)]%
        {yazdinejad2021review}
\bibfield{author}{\bibinfo{person}{Abbas Yazdinejad}, \bibinfo{person}{Behrouz
  Zolfaghari}, \bibinfo{person}{Amin Azmoodeh}, \bibinfo{person}{Ali
  Dehghantanha}, \bibinfo{person}{Hadis Karimipour}, \bibinfo{person}{Evan
  Fraser}, \bibinfo{person}{Arthur~G Green}, \bibinfo{person}{Conor Russell},
  {and} \bibinfo{person}{Emily Duncan}.} \bibinfo{year}{2021}\natexlab{}.
\newblock \showarticletitle{A review on security of smart farming and precision
  agriculture: Security aspects, attacks, threats and countermeasures}.
\newblock \bibinfo{journal}{\emph{Applied Sciences}} \bibinfo{volume}{11},
  \bibinfo{number}{16} (\bibinfo{year}{2021}), \bibinfo{pages}{7518}.
\newblock


\bibitem[Yun et~al\mbox{.}(2022)]%
        {yun2022cooperative}
\bibfield{author}{\bibinfo{person}{Won~Joon Yun}, \bibinfo{person}{Soohyun
  Park}, \bibinfo{person}{Joongheon Kim}, \bibinfo{person}{MyungJae Shin},
  \bibinfo{person}{Soyi Jung}, \bibinfo{person}{David~A Mohaisen}, {and}
  \bibinfo{person}{Jae-Hyun Kim}.} \bibinfo{year}{2022}\natexlab{}.
\newblock \showarticletitle{Cooperative multiagent deep reinforcement learning
  for reliable surveillance via autonomous multi-{UAV} control}.
\newblock \bibinfo{journal}{\emph{IEEE Trans. Industrial Informatics}}
  \bibinfo{volume}{18}, \bibinfo{number}{10} (\bibinfo{year}{2022}),
  \bibinfo{pages}{7086--7096}.
\newblock


\bibitem[Zhu et~al\mbox{.}(2023)]%
        {zhu2023transfer}
\bibfield{author}{\bibinfo{person}{Zhuangdi Zhu}, \bibinfo{person}{Kaixiang
  Lin}, \bibinfo{person}{Anil~K. Jain}, {and} \bibinfo{person}{Jiayu Zhou}.}
  \bibinfo{year}{2023}\natexlab{}.
\newblock \showarticletitle{Transfer Learning in Deep Reinforcement Learning: A
  Survey}.
\newblock \bibinfo{journal}{\emph{IEEE Trans. Pattern Analysis and Machine
  Intelligence}} \bibinfo{volume}{45}, \bibinfo{number}{11}
  (\bibinfo{year}{2023}), \bibinfo{pages}{13344--13362}.
\newblock


\end{thebibliography}

\end{document}